# A Comprehensive Capability Analysis of GPT-3 and GPT-3.5 Series Models

Junjie Ye[★][*], Xuanting Chen[★][*], Nuo Xu[★], Can Zu[★], Zekai Shao[♠],
Shichun Liu[★], Yuhan Cui[★], Zeyang Zhou[★], Chao Gong[★], Yang Shen[★],
Jie Zhou[★], Siming Chen[♠], Tao Gui[♦][†], Qi Zhang[★][♣], Xuanjing Huang[★]

[★] School of Computer Science, Fudan University, Shanghai, China
[♠] School of Data Science, Fudan University, Shanghai, China
[♦] Institute of Modern Languages and Linguistics, Fudan University, Shanghai, China
[♣] Shanghai Collaborative Innovation Center of Intelligent Visual Computing, Fudan University
{jjye19,tgui,qz,xjhuang}@fudan.edu.cn
xuantingchen21@m.fudan.edu.cn

## Abstract

GPT series models, such as GPT-3, CodeX, InstructGPT, ChatGPT, and so on, have gained considerable attention due to their exceptional natural language processing capabilities. However, despite the abundance of research on the difference in capabilities between GPT series models and fine-tuned models, there has been limited attention given to the evolution of GPT series models' capabilities over time. To conduct a comprehensive analysis of the capabilities of GPT series models, we select six representative models, comprising two GPT-3 series models (i.e., davinci and text-davinci-001) and four GPT-3.5 series models (i.e., code-davinci-002, text-davinci-002, text-davinci-003, and gpt-3.5-turbo). We evaluate their performance on nine natural language understanding (NLU) tasks using 21 datasets. In particular, we compare the performance and robustness of different models for each task under zero-shot and few-shot scenarios. Our extensive experiments reveal that the overall ability of GPT series models on NLU tasks does not increase gradually as the models evolve, especially with the introduction of the RLHF training strategy. While this strategy enhances the models' ability to generate human-like responses, it also compromises their ability to solve some tasks. Furthermore, our findings indicate that there is still room for improvement in areas such as model robustness.

## 1 Introduction

Large language models (LLMs), such as FLAN (Wei et al., 2022), OPT (Zhang et al., 2022b), and PaLM (Chowdhery et al., 2022), have demonstrated exceptional performance in natural language understanding (NLU) tasks. Among these models, the Generative Pre-trained Transformer (GPT) (Brown et al., 2020) series has recently garnered significant interest due to their outstanding performance in unifying all NLU tasks into generative tasks. Specifically, the GPT series models comprise two sub-series: GPT-3 and GPT-3.5, with their evolutionary relationship depicted in Figure 1, as documented by OpenAI [1].

Extensive research has been conducted to explore the capabilities of these models from various perspectives. On one hand, researchers have performed experiments to evaluate the performance of GPT series models in specific natural language processing (NLP) tasks. For instance, Zhang et al. (2022a) demonstrated that GPT-3 has acquired linguistic knowledge and can recognize semantic information in most continuous contexts. In addition, Yang et al. (2023) and Hendy et al. (2023) investigated the potential of ChatGPT (i.e., gpt-3.5-turbo in Figure 1) in ascpect-based text summarization and machine translation tasks, respectively. Furthermore, (Qin et al., 2023) analyzed the zero-shot capability of ChatGPT across seven representative task categories. On the other hand, some researchers have investigated the limitations of GPT series models. For example, Koco'n et al. (2023) compared the performance of ChatGPT with state-of-the-art models on 25 different NLP tasks, revealing certain biases and shortcomings of ChatGPT. Additionally, Chen et al. (2023) conducted robustness tests on the GPT series models on 9 NLU tasks and demonstrated that these models still experience similar problems with robustness as fine-tuned models.

---

[*] Equal contribution.
[†] Corresponding author.
[1] https://platform.openai.com/docs/model-index-for-researchers

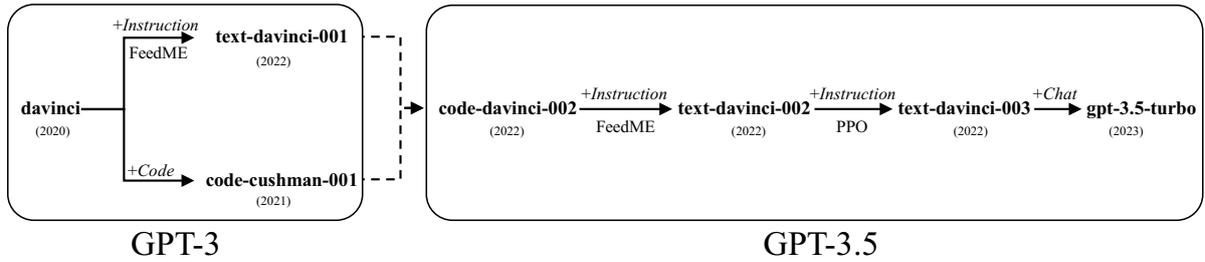

Figure 1: The evolutionary relationship of the GPT series models. FeedME and PPO are two distinct training strategies officially described by OpenAI. A dashed arrow (--→) is used between GPT-3 and GPT-3.5 since the official documentation does not provide specific information on the differences between the two series when trained.

However, while many studies have focused on comparing the performance of specific GPT series models to fine-tuned models for particular tasks or analyzing their shortcomings relative to fine-tuned models, a comprehensive analysis of the evolution of GPT series models is still lacking. Specifically, there is a need to investigate how the different strategies used in training GPT series models impact their capabilities in NLU tasks.

In order to conduct a comprehensive analysis of the capabilities of the GPT-3 and GPT-3.5 series models, we evaluate the performance of six GPT series models across nine different NLU tasks using 21 datasets and corresponding transformation data generated by TextFlint (Gui et al., 2021). These models include two GPT-3 series models (i.e., davinci and text-davinci-001) and four GPT-3.5 series models (i.e., code-davinci-002, text-davinci-002, text-davinci-003, and gpt-3.5-turbo). Our analysis focused on three main perspectives: 1) comparing the performance of different models across various NLU tasks; 2) examining the effect of the training strategies employed by the GPT series models on their capabilities; and 3) analyzing the effect of zero-shot and few-shot scenarios on the capabilities of the models.

Our *findings* are summarized as follows:

- **Davinci lacks instruction comprehension.** The davinci model cannot produce an answer in the zero-shot scenario for prompts that are declarative sentences and do not end with a word such as "Answer", indicating a lack of instruction comprehension (Section 4.1.2).

- **In-context learning improves prompt understanding for davinci.** For the davinci model, in the named entity recognition (NER) and part-of-speech (POS) tasks, in-context learning substantially improves the proportion of outputs that meet the instruction requirements in the few-shot scenario, while in the inference-based tasks (e.g., natural language inference (NLI), relation extraction (RE) and the winograd scheme challenge (WSC)), in-context learning does not significantly improve performance, suggesting its usefulness in helping the model understand prompts (Section 4.1.1).

- **All models are sensitive to prompts.** We select three prompts for each task in different scenarios to test the ability of the models other than davinci [2], and the results show that all models exhibit prompt sensitivity in both zero-shot and few-shot scenarios, but the extent of sensitivity varies across different models and tasks and requires further investigation (Section 4.2).

- **Different models perform differently in zero-shot scenarios.** In the zero-shot scenario, code-davinci-002 performs best in aspect-based sentiment analysis (ABSA), machine reading comprehension (MRC), and sentiment classification (SC) tasks; text-davinci-003 performs best in POS, RE, and semantic matching (SM) tasks; gpt-3.5-turbo performs better in NLI and wSC tasks, but has difficulty following instructions in the POS task, which is similar to that of text-davinci-001.

---

[2]The prompts are listed in Appendix B.

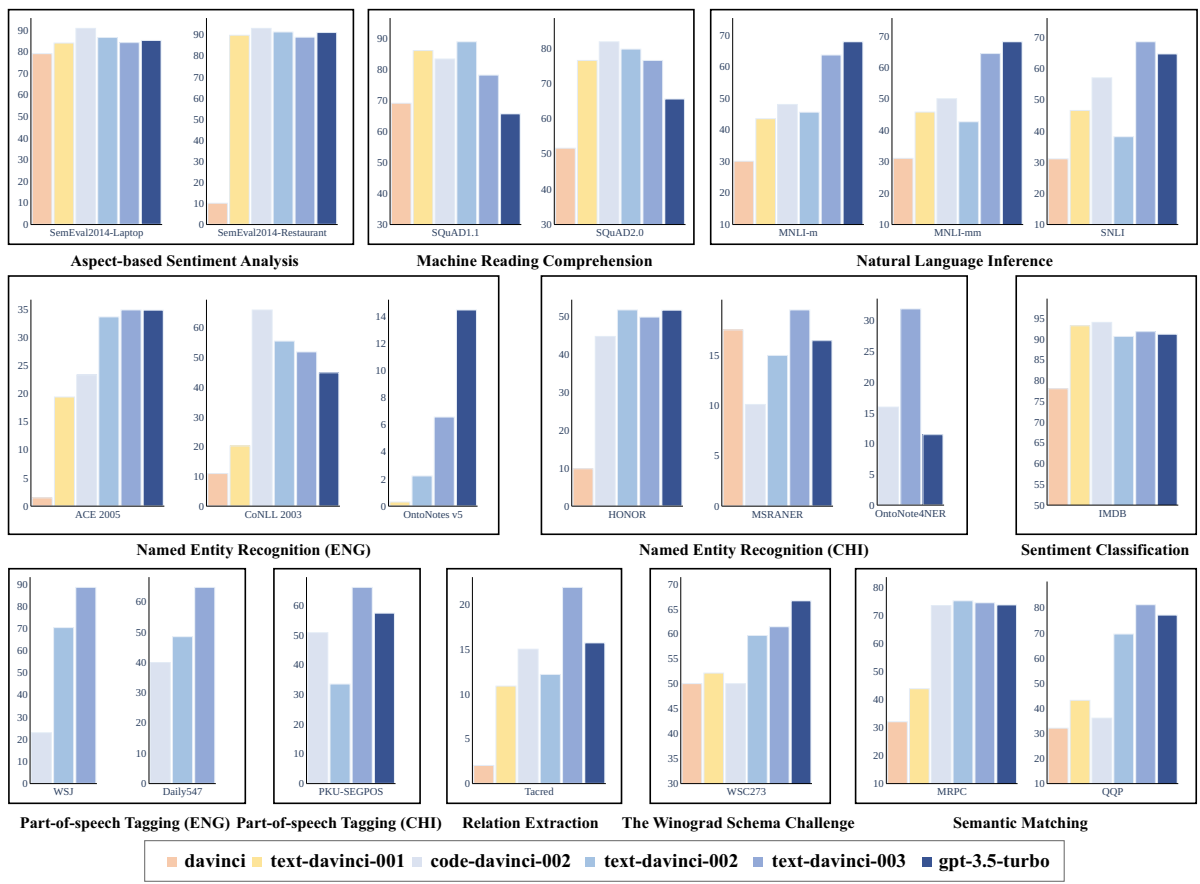

Figure 2: The performance of different models in zero-shot scenario. Missing bars in some datasets mean that the model cannot perform the specified task on that dataset. See Appendix A.1 for specific data.

This may be due to gpt-3.5-turbo using a smaller model and weakening the ability of tasks where interaction with humans is not important (Section 4.2 and Figure 2).

- **Few-shot scenarios do not always improve model performance.** Although models generally perform better in the few-shot scenario than in the zero-shot scenario, this is not always the case and depends on the model, task, prompt design, and example selection, which deserves further study. (Section 4.2).

- **Text-davinci-001 has relatively weak capabilities compared to other models.** Compared to other models except davinci, text-davinci-001 has the weakest overall ability on most tasks, but still showed moderate performance in two tasks, MRC and SC (Section 4.2).

- **Gpt-3.5-turbo and text-davinci-003 have comparable capabilities.** Compared to text-davinci-003, gpt-3.5-turbo has similar performance to it on most tasks, and only has a disadvantage in MRC, POS, and RE tasks, which may be due to its smaller model size. This of course needs to be studied in more depth (Section 4.2).

- **Increasing model capability does not always improve robustness.** With the exception of the ABSA task, where different models show some differences in robustness, the robustness of different models in other tasks is relatively similar, indicating that there is still much room for improvement in model robustness (Section 4.2).

Based on these findings, we draw the following *conclusions*:

- **The pre-training phase provides the model with fundamental comprehension and in-context learning abilities.** For example, the davinci model is a text generation model that does not require explicit instructions during pre-training. However, even in the zero-shot scenario, it can understand task instructions for tasks including NLI, SC, SM, WSC, and generate effective answers. In the few-shot scenario, the model's understanding of instructions for complex tasks like NER and POS is greatly improved, leading to more analyzable answers (Section 4.1.1).

- **The inclusion of a certain type of task in the supervised fine-tuning phase may have a significant impact on the model's performance on that type of task.** For instance, while text-davinci-001 performs poorly on NER and POS tasks, it shows similar performance to text-davinci-002 and text-davinci-003 on the MRC task. However, since we cannot determine from official documentation which tasks the model uses for supervised fine-tuning, this issue warrants further investigation (Section 4.2).

- **Alignment with human cognition to some extent impairs the performance of the model on certain tasks.** Text-davinci-002, an InstructGPT model based on code-davinci-002, exhibits performance advantages over the latter in SM and WSC tasks, but its performance on other tasks is similar or even worse than code-davinci-002, particularly in few-shot scenarios. OpenAI refers to this phenomenon as the "alignment tax" (Ouyang et al., 2022) (Figure 1 and Section 4.2).

- **RLHF (Christiano et al., 2017) is leveraged to enhance the model's ability to produce human-like responses, rather than directly improving its performance.** Text-davinci-003 is an improvement over text-davinci-002, as it incorporates RLHF as a training strategy. However, its performance is comparable to that of text-davinci-002 in most tasks and inferior to text-davinci-002 in SC and SM tasks. This is due to the fact that RLHF provides limited knowledge to support the model's deeper understanding of the task, thereby not significantly improving the model's performance in NLU tasks (Figure 1 and Section 4.2).

## 2 Background

### 2.1 GPT-3 and GPT-3.5 Series Models

GPT-3 and GPT-3.5 are a series of language models developed by OpenAI for generating human-like natural language text. As depicted in Figure 1, davinci is the basis of the GPT-3 series of models and has 175 billion parameters, making it a highly capable text generator. OpenAI has pursued two upgrade paths for davinci: supervised fine-tuning training to create InstructGPT (Ouyang et al., 2022), text-davinci-001, and code training to create Codex (Chen et al., 2021), code-cushman-001. In 2022, OpenAI released code-davinci-002 for code generation tasks, which became the base model for the GPT-3.5 series. OpenAI then used supervised fine-tuning to create text-davinci-002 and introduced the RLHF training strategy to create text-davinci-003, which improved its ability to understand instructions and generate text. Based on text-davinci-003, OpenAI optimized gpt-3.5-turbo for chat, which is the most capable GPT-3.5 model available at a lower cost than text-davinci-003 [3]. In this paper, we conduct an extensive comparative analysis experiment of GPT-3 and GPT-3.5 series models to explain their evolution and the impact of different training strategies on their capabilities.

### 2.2 TextFlint

TextFlint is a multilingual platform for evaluating the robustness of NLP tasks. It provides comprehensive analysis by integrating general and task-specific text transformations, adversarial attacks, subgroups, and combinations thereof. TextFlint uses a custom production-to-analysis workflow to address challenges related to completeness, acceptability, and analyzability. The platform offers over 80 data transformation methods designed for 12 NLP tasks, including 20 general and 60 domain-specific transformations.In this paper, 16 of the 21 datasets we used were collated by TextFlint.

---

[3] https://platform.openai.com/docs/models/gpt-3-5

Table 1: Information of all datasets used in experiments.

| Task | Dataset | # Samples | Measure | Language |
|---|---|---|---|---|
| Aspect-based Sentiment Analysis | SemEval2014-Laptop | 331 | Accuracy | English |
|  | SemEval2014-Restaurant | 492 | Accuracy | English |
| Machine Reading Comprehension | SQuAD1.1 | 9868 | F1 & EM | English |
|  | SQuAD2.0 | 11491 | F1 & EM | English |
| Named Entity Recognition | ACE 2005 | 1312 | F1 | English |
|  | CoNLL 2003 | 3453 | F1 | English |
|  | OntoNotes v5 | 4019 | F1 | English |
|  | HONOR | 1120 | F1 | Chinese |
|  | MSRANER | 4365 | F1 | Chinese |
|  | OntoNote4NER | 4346 | F1 | Chinese |
| Natural Language Inference | MNLI-m | 9815 | Accuracy | English |
|  | MNLI-mm | 9832 | Accuracy | English |
|  | SNLI | 10000 | Accuracy | English |
| Part-of-speech Tagging | Daily547 | 546 | Accuracy | English |
|  | WSJ | 5461 | Accuracy | English |
|  | PKU-SEGPOS | 5204 | F1 | Chinese |
| Relation Extraction | Tacred | 15509 | F1 | English |
| Sentiment Classification | IMDB | 11257 | Accuracy | English |
| Semantic Matching | MRPC | 1724 | Accuracy | English |
|  | QQP | 5000 | Accuracy | English |
| The Winograd Schema Challenge | WSC273 | 570 | Accuracy | English |

## 3 Experiment Setup

### 3.1 Datasets

We conduct an evaluation of the capabilities of GPT-3 and GPT-3.5 series models, covering 9 different NLU tasks using 21 datasets and corresponding transformation data generated by TextFlint: **ABSA** (SemEval2014-Laptop (Pontiki et al., 2014) and SemEval2014-Restaurant (Pontiki et al., 2014)), **MRC** (SQuAD1.1 (Rajpurkar et al., 2016) and SQuAD2.0 (Rajpurkar et al., 2018)), **NER** (ACE2005 [4], CoNLL2003 (Sang and Meulder, 2003), OntoNotesv5 [5], HONOR (Chen et al., 2023), MSRANER (Levow, 2006), and OntoNote4NER (Weischedel et al., 2013)), **NLI** (MNLI-m (Williams et al., 2017), MNLI-mm (Williams et al., 2017), and SNLI (Williams et al., 2017)), **POS** (WSJ (Marcus et al., 1993), Daily547 (Gimpel et al., 2010), and PKU-SEGPOS [6]), **RE** (Tacred (Zhang et al., 2017)), **SC** (IMDB (Maas et al., 2011)), **SM** (MRPC (Dolan and Brockett, 2005), QQP (Wang et al., 2017)), and **WSC** (WSC273 (Levesque et al., 2012)). The information of different datasets is shown in Table 1.

### 3.2 GPT Systems

We have selected six GPT series models to represent their evolution, all of which are evaluated using OpenAI's official API [7]:

---

[4] https://catalog.ldc.upenn.edu/LDC2006T06
[5] https://catalog.ldc.upenn.edu/LDC2013T19
[6] http://cuge.baai.ac.cn/#/dataset?id=19&name=PKU-SEGPOS
[7] https://platform.openai.com

- **Davinci**: the base of GPT-3 series models, which can can understand and generate natural language with higher quality.

- **Text-davinci-001**: an InstructGPT model based on davinci, using Feedback Made Easy (FeedME) strategy, which involves supervised fine-tuning on human-written demonstrations and on model samples rated 7/7 by human labelers on an overall quality score [8].

- **Code-davinci-002**: the most capable Codex model, which is a descendant of GPT-3 and the base of GPT-3.5 series models, with training data that contains both natural language and billions of lines of source code from publicly available sources, including code in public GitHub repositories.

- **Text-davinci-002**: an InstructGPT model based on code-davinci-002, trained with FeedME method.

- **Text-davinci-003**: an improvement version of text-davinci-002, but trained with Proximal Policy Optimization (PPO) algorithm, which is used in reinforcement learning with reward models trained from comparisons by humans Ouyang et al. (2022).

- **Gpt-3.5-turbo**: the most capable GPT-3.5 model and optimized for chat at 1/10th the cost of text-davinci-003.

Please note that when evaluating davinci and code-davinci-002, we test on 100 samples, while for text-davinci-001 and text-davinci-002 on 1000 samples [9]. For text-davinci-003 and gpt-3.5-turbo, we use the entire dataset for evaluation.

### 3.3 Prompt Selection Strategies

LLMs have shown promising results through in-context learning by using a few labeled examples, known as prompts, in addition to the test input. This approach, known as the few-shot paradigm, has demonstrated good performance across multiple NLP tasks.

In this paper, we gather a large number of task-specific prompts from various sources, including the GitHub repository "promptsource" [10]. And we manually design new prompts for specific tasks, selecting the three best-performing prompts per dataset to ensure the most objective results. We expand these prompts into both zero-shot and few-shot scenarios by varying the number of examples in the prompt. We also map the original labels to specific phrases in the prompt for the RE, NER, and POS tasks to help the model understand their meaning. More information on these prompts can be found in Appendix B.

---

[8] https://platform.openai.com/docs/model-index-for-researchers
[9] For datasets with less than 1000 records, we choose the full amount of data.
[10] https://github.com/bigscience-workshop/promptsource

# 4 Experiments

## 4.1 Performance of Davinci

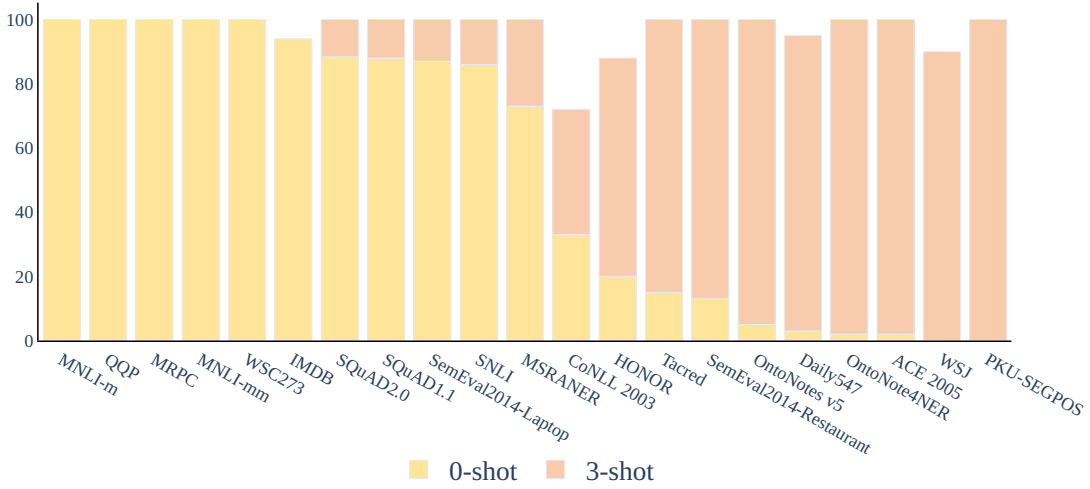

Figure 3: The analyzability rates of davinci's performance on different datasets in both zero-shot and three-shot scenarios, with the results ordered based on the ratio of three-shot to zero-shot performance. The details of results are listed in Appendix A.2.

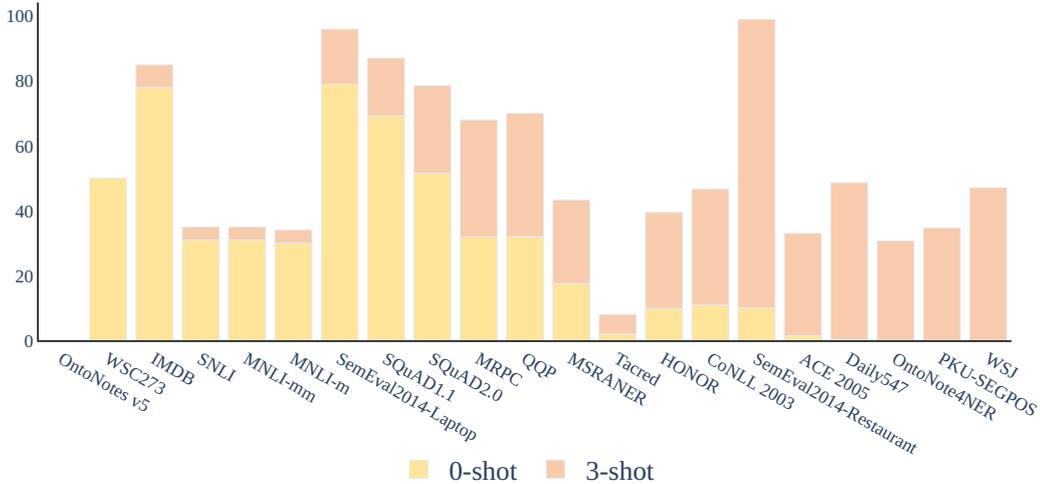

Figure 4: The performance of davinci on different datasets in both zero-shot and three-shot scenarios, with the results ordered based on the ratio of three-shot to zero-shot performance. The details of results are listed in Appendix A.2.

We evaluate the performance of the davinci model in both zero-shot and three-shot scenarios across different datasets. We report the analyzable rate and corresponding performance results in Figure 3 and Figure 4. From the figures, it is evident that davinci exhibits good analyzability and achieves good performance on many datasets (e.g., MNLI-m, MNLI-mm, IMDB, and WSC273) even in the ZERO-shot case, without the use of supervised fine-tuning. For the datasets where zero-shot performance is not possible (e.g., ACE 2005, WSJ, and PKU-SEGPOS), davinci effectively learn from the examples provided in the three-shot scenario. This demonstrates that the pre-training phase equips the model with basic understanding and in-context learning abilities.

### 4.1.2 Instruction Comprehension of davinci

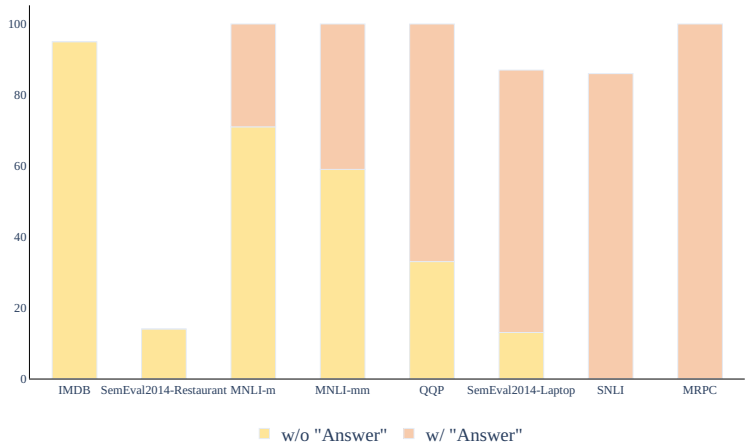

Figure 5: The analyzability rates of davinci's answer results in the zero-shot scenario. The results are ordered based on the ratio of the analyzability rate when the prompt includes the word "Answer" at the end, to the rate when it does not. The details of results are listed in Appendix A.3.

In the zero-shot scenario, we choose some datasets to test davinci's instruction comprehension and obtain Figure 5. Specifically, we remove the "Answer" at the end of the prompt, which is present in the tests in Section 4.1.1. Surprisingly, when the word "Answer" is removed from the prompt, the analyzability of davinci's results on most of the datasets drops severely, if at all, to produce an answer. This illustrates the lack of instruction comprehension in davinci, and therefore, the inclusion of instructions in training is necessary for the model to handle NLU problems.

### 4.2 Comparison Experiments

#### 4.2.1 Aspect-based Sentiment Analysis

Table 2: Performance and robustness test results (accuracy) of GPT series models in zero-shot and few-shot scenarios on **SemEval2014-Laptop** dataset.

| Model | AddDiff # 331 samples | | ReverseNonTarget # 104 samples | | ReverseTarget # 331 samples | |
|---|---|---|---|---|---|---|
| | ori | trans | ori | trans | ori | trans |
| *0-shot* | | | | | | |
| code-davinci-002 | **92.88±2.14** | **90.18±7.42** | **91.39±2.96** | **53.09±2.78** | **93.23±1.65** | **58.61±0.89** |
| text-davinci-001 | 85.21±1.70 | 80.10±2.11 | 85.89±1.68 | 47.35±4.16 | 85.26±2.17 | 53.56±0.92 |
| text-davinci-002 | 86.38±0.11 | 81.90±0.35 | 85.57±0.21 | 52.97±2.96 | 86.40±0.26 | 56.68±5.17 |
| text-davinci-003 | 83.84±0.33 | 77.50±2.43 | 82.43±0.42 | 39.61±4.83 | 83.62±0.12 | 47.04±4.64 |
| gpt-3.5-turbo | 85.57±1.27 | 86.55±8.67 | 88.78±2.22 | 41.78±7.36 | 85.67±1.36 | 49.75±9.51 |
| *1-shot* | | | | | | |
| code-davinci-002 | 96.33±0.58 | 92.67±0.58 | 94.00±1.00 | 53.33±1.53 | 96.33±0.58 | **65.00±1.00** |
| text-davinci-001 | 82.87±1.04 | 72.69±0.63 | 82.85±1.68 | 45.74±2.29 | 82.94±0.88 | 47.50±3.15 |
| text-davinci-002 | 86.05±0.43 | 82.20±1.98 | 85.22±0.24 | 55.18±2.38 | 86.05±0.43 | 56.77±3.08 |
| text-davinci-003 | 85.77±0.69 | 87.17±4.62 | 85.63±1.05 | 52.22±8.86 | 85.84±0.57 | 57.07±7.43 |
| gpt-3.5-turbo | 88.99±0.73 | 85.17±4.90 | 93.22±1.00 | 41.93±5.21 | 89.18±0.90 | 47.95±6.93 |
| *3-shot* | | | | | | |
| code-davinci-002 | **97.00±1.00** | **94.00±1.00** | 94.00±0.00 | 52.00±2.65 | **97.00±1.00** | 64.33±1.53 |
| text-davinci-001 | 83.33±0.69 | 71.90±0.12 | 83.40±0.24 | 48.40±1.98 | 83.44±0.87 | 50.26±0.77 |
| text-davinci-002 | 85.41±0.43 | 81.55±1.86 | 84.80±0.97 | 54.01±2.28 | 85.48±0.33 | 56.65±2.43 |
| text-davinci-003 | 85.91±0.12 | 88.73±4.11 | 85.08±0.48 | **55.59±11.11** | 85.98±0.12 | 59.14±7.11 |
| gpt-3.5-turbo | 90.75±1.57 | 90.23±6.40 | 93.09±1.98 | 47.93±5.83 | 90.45±1.08 | 53.62±4.82 |

Table 3: Performance and robustness test results (accuracy) of GPT series models in zero-shot and few-shot scenarios on **SemEval2014-Restaurant** dataset.

| Model | AddDiff # 492 samples | | ReverseNonTarget # 227 samples | | ReverseTarget # 492 samples | |
|---|---|---|---|---|---|---|
| | ori | trans | ori | trans | ori | trans |
| *0-shot* | | | | | | |
| code-davinci-002 | **94.65±2.09** | 57.23±29.53 | **97.00±2.00** | 74.33±3.51 | **94.31±2.33** | 72.92±5.25 |
| text-davinci-001 | 89.25±0.94 | 54.56±12.55 | 90.07±1.24 | 63.35±1.58 | 88.89±1.45 | 63.40±1.94 |
| text-davinci-002 | 91.38±0.21 | **70.41±13.90** | 92.52±0.26 | 66.26±1.66 | 91.52±0.33 | 68.68±3.02 |
| text-davinci-003 | 89.45±0.87 | 55.25±20.03 | 91.58±0.96 | 47.93±0.45 | 89.38±0.85 | 54.60±7.18 |
| gpt-3.5-turbo | 90.70±0.41 | 64.21±18.97 | 92.64±1.54 | 70.40±6.69 | 90.94±0.47 | 63.39±7.76 |
| *1-shot* | | | | | | |
| code-davinci-002 | 98.00±1.00 | 75.67±14.64 | **100.00±0.00** | 78.67±4.16 | 98.00±1.00 | **78.67±4.51** |
| text-davinci-001 | 88.07±3.09 | 33.70±4.97 | 88.49±3.02 | 61.90±1.22 | 88.03±3.05 | 58.03±3.50 |
| text-davinci-002 | 91.86±0.79 | 74.05±19.15 | 92.00±0.91 | 69.50±3.12 | 91.90±0.79 | 69.96±5.18 |
| text-davinci-003 | 92.08±0.74 | 78.30±17.09 | 92.84±0.52 | 64.97±11.58 | 92.05±0.78 | 66.62±8.78 |
| gpt-3.5-turbo | 92.34±1.11 | 60.23±20.70 | 94.42±2.08 | 67.79±6.21 | 92.34±1.11 | 60.45±5.68 |
| *3-shot* | | | | | | |
| code-davinci-002 | **98.00±0.00** | 84.67±9.29 | **100.00±0.00** | 76.33±2.89 | **98.00±0.00** | 74.00±3.00 |
| text-davinci-001 | 89.49±0.21 | 50.57±3.50 | 90.43±0.52 | 62.07±1.04 | 89.57±0.25 | 60.73±1.02 |
| text-davinci-002 | 92.48±0.54 | 86.46±7.66 | 92.39±0.48 | 68.94±2.88 | 92.41±0.61 | 72.39±4.25 |
| text-davinci-003 | 92.56±0.83 | **89.47±7.53** | 93.01±0.52 | 70.95±5.76 | 92.52±0.78 | 70.68±4.60 |
| gpt-3.5-turbo | 94.62±0.12 | 69.98±16.42 | 96.31±0.29 | 71.75±5.24 | 94.55±0.13 | 65.18±2.83 |

We analyze the performance of various models on two ABSA datasets, namely SemEval2014-Laptop and SemEval2014-Restaurant, and present the outcomes in Table 2 and 3. While some models demonstrate good performance on these datasets, there are still issues with robustness. Our analysis comprises two scenarios: zero-shot and few-shot, and the tables display the results.

**In the zero-shot scenario, all models' performance on the ABSA task is nearly identical, presumably because it is a simpler task.** Specifically, code-davinci-002 exhibits the most consistent performance, followed by gpt-3.5-turbo and text-davinci-002. Text-davinci-001's performance is poor in most tasks but relatively better in the ABSA task. All five models demonstrate poor performance in other variations except for "AddDiff", particularly the davinci series models.

**In the few-shot scenario, code-davinci-002 demonstrates further enhancement relative to the zero-shot scenario, achieving zero errors in the "ReverseNonTarget" variation of SemEval2014-Restaurant.** The five models' performance in zero-shot and few-shot is comparable, indicating that these two datasets are not significantly influenced by the number of examples in the prompt. Concerning robustness, the GPT series models do not demonstrate any significant changes with iterative updates.

### 4.2.2 Machine Reading Comprehension

Table 4: Performance and robustness test results (micro-F1) of GPT series models in zero-shot and few-shot scenarios on **SQuAD1.1** dataset.

| Model | AddSentDiverse # 9292 samples | | ModifyPos # 9011 samples | | PerturbAnswer # 9833 samples | | PerturbQuestion-BackTranslation # 9868 samples | | PerturbQuestion-MLM # 9867 samples | |
|---|---|---|---|---|---|---|---|---|---|---|
| | ori | trans | ori | trans | ori | trans | ori | trans | ori | trans |
| *0-shot* | | | | | | | | | | |
| code-davinci-002 | **83.58±2.72** | **66.79±2.14** | 82.32±2.39 | 83.64±0.82 | 84.10±2.23 | 82.67±3.00 | 83.06±2.59 | 71.06±3.22 | 83.30±2.73 | 61.79±7.76 |
| text-davinci-001 | 73.38±11.70 | 51.64±12.22 | 72.45±11.93 | 72.53±11.08 | 72.93±11.72 | 65.89±10.78 | 73.17±11.69 | 62.47±9.26 | 73.09±11.62 | 61.27±9.91 |
| text-davinci-002 | 78.56±11.01 | 55.67±8.01 | 77.77±11.76 | 77.10±11.19 | 78.52±10.86 | 72.21±10.85 | 78.44±10.84 | 69.04±9.52 | 78.47±10.94 | **64.47±10.43** |
| text-davinci-003 | 66.84±9.79 | 55.59±9.95 | 67.55±9.84 | 67.46±8.94 | 67.16±9.71 | 65.97±8.27 | 67.19±9.66 | 59.90±8.00 | 67.18±9.64 | 56.43±9.15 |
| gpt-3.5-turbo | 55.26±10.08 | 37.33±5.70 | 56.71±10.37 | 55.67±9.30 | 54.87±9.03 | 47.21±7.48 | 54.86±9.40 | 46.74±7.58 | 54.95±9.53 | 35.81±5.86 |
| *1-shot* | | | | | | | | | | |
| code-davinci-002 | 86.62±0.71 | **75.52±2.18** | 85.84±0.79 | 87.68±1.74 | 86.24±0.39 | **87.96±1.99** | 86.36±0.25 | 79.74±0.84 | 86.52±0.74 | 81.79±0.60 |
| text-davinci-001 | 85.01±2.11 | 67.67±0.97 | 84.83±1.74 | 84.28±2.68 | 85.00±2.03 | 79.28±2.11 | 85.11±1.90 | 73.43±1.54 | 85.02±1.90 | 73.02±1.36 |
| text-davinci-002 | 58.82±19.61 | 47.43±14.23 | 58.19±19.65 | 57.48±19.94 | 58.42±19.04 | 52.35±15.05 | 58.62±19.10 | 52.43±16.37 | 58.30±18.93 | 49.55±18.01 |
| text-davinci-003 | 88.20±1.30 | 70.75±1.98 | 88.47±1.24 | 88.37±1.00 | 88.13±1.27 | 85.27±1.11 | 88.10±1.29 | 80.31±1.19 | 88.13±1.26 | 78.34±1.70 |
| gpt-3.5-turbo | 80.31±2.37 | 58.36±4.12 | 81.36±2.24 | 79.97±2.11 | 79.34±2.63 | 70.47±2.51 | 79.52±2.30 | 67.24±1.95 | 79.14±2.96 | 52.64±1.86 |
| *3-shot* | | | | | | | | | | |
| code-davinci-002 | 86.55±0.41 | 67.65±3.27 | 85.38±1.34 | 85.98±1.85 | 86.69±0.93 | 87.91±2.36 | 86.74±0.33 | 77.80±1.47 | 86.61±0.97 | 83.53±2.24 |
| text-davinci-001 | 85.41±0.87 | 65.86±1.10 | 85.49±1.31 | 85.33±2.21 | 85.73±0.39 | 80.17±0.61 | 85.30±0.95 | 74.28±0.97 | 85.37±0.85 | 74.81±0.68 |
| text-davinci-002 | 82.72±7.87 | 61.97±2.90 | 83.37±8.13 | 82.65±8.01 | 82.80±7.74 | 80.98±7.24 | 82.93±7.66 | 74.96±7.82 | 82.93±7.70 | 71.64±10.81 |
| text-davinci-003 | **89.64±0.40** | 68.04±0.85 | **89.94±0.35** | 89.88±0.38 | 89.60±0.37 | 86.92±0.44 | 89.57±0.38 | 82.10±0.55 | 89.57±0.38 | 79.94±0.90 |
| gpt-3.5-turbo | 80.65±0.63 | 57.99±1.00 | 81.92±0.46 | 80.91±0.48 | 80.13±0.21 | 71.36±0.64 | 80.03±0.56 | 66.83±0.93 | 79.86±0.31 | 51.76±0.67 |

Table 5: Performance and robustness test results (Exact Match) of GPT series models in zero-shot and few-shot scenarios on **SQuAD1.1** dataset.

| Model | AddSentDiverse # 9292 samples | | ModifyPos # 9011 samples | | PerturbAnswer # 9833 samples | | PerturbQuestion-BackTranslation # 9868 samples | | PerturbQuestion-MLM # 9867 samples | |
|---|---|---|---|---|---|---|---|---|---|---|
| | ori | trans | ori | trans | ori | trans | ori | trans | ori | trans |
| *0-shot* | | | | | | | | | | |
| code-davinci-002 | **78.35±2.06** | **60.00±2.65** | 76.35±2.05 | 76.33±1.53 | 78.33±2.08 | 76.94±4.91 | 77.29±2.15 | 61.33±2.52 | 77.62±2.32 | 54.50±8.78 |
| text-davinci-001 | 58.17±16.43 | 38.40±16.18 | 56.67±16.95 | 57.07±15.46 | 57.63±16.28 | 50.03±15.24 | 57.87±16.34 | 47.87±13.25 | 57.73±16.09 | 47.57±13.37 |
| text-davinci-002 | 65.33±15.96 | 43.13±12.76 | 63.57±17.25 | 62.80±16.62 | 65.27±15.61 | 58.07±15.88 | 65.33±15.61 | 55.27±13.57 | 65.30±15.70 | 50.50±13.65 |
| text-davinci-003 | 42.76±15.43 | 34.01±15.57 | 43.64±15.58 | 43.71±14.66 | 43.37±15.34 | 42.58±13.32 | 43.37±15.31 | 36.73±12.78 | 43.34±15.31 | 34.05±13.79 |
| gpt-3.5-turbo | 27.93±14.11 | 12.19±7.68 | 28.94±14.98 | 27.45±13.24 | 26.77±12.57 | 20.41±10.07 | 26.68±13.29 | 21.17±10.96 | 26.74±13.37 | 15.07±7.63 |
| *1-shot* | | | | | | | | | | |
| code-davinci-002 | 77.82±15.81 | **77.25±20.62** | 76.80±15.16 | 71.85±30.77 | 81.87±13.67 | 85.54±4.00 | 83.48±11.91 | 69.95±28.75 | 74.52±9.80 | 69.50±25.38 |
| text-davinci-001 | 73.43±3.39 | 57.43±0.76 | 73.27±2.85 | 73.17±4.00 | 73.43±3.20 | 66.97±2.52 | 73.50±3.02 | 62.07±2.65 | 73.50±3.06 | 62.00±2.26 |
| text-davinci-002 | 51.20±19.81 | 40.97±14.12 | 50.53±19.99 | 49.87±20.40 | 50.63±19.27 | 44.57±14.77 | 50.90±19.33 | 44.50±15.96 | 50.47±19.17 | 41.27±18.21 |
| text-davinci-003 | 74.51±3.13 | 57.93±3.08 | 75.03±3.00 | 75.26±2.60 | 74.55±3.07 | 70.66±2.65 | 74.47±3.07 | 65.54±2.80 | 74.50±3.04 | 62.86±3.14 |
| gpt-3.5-turbo | 62.05±4.23 | 40.96±5.16 | 63.62±4.15 | 62.02±4.04 | 60.01±4.44 | 51.85±4.40 | 60.16±4.06 | 48.90±3.61 | 59.54±4.91 | 36.44±3.58 |
| *3-shot* | | | | | | | | | | |
| code-davinci-002 | **86.55±0.41** | 67.65±3.27 | **85.38±1.34** | 85.98±1.85 | **86.69±0.93** | 87.91±2.36 | 86.74±0.33 | 77.80±1.47 | 86.61±0.97 | 83.53±2.24 |
| text-davinci-001 | 76.05±1.41 | 57.78±1.23 | 76.16±2.05 | 76.23±3.42 | 76.50±1.37 | 70.35±1.33 | 75.97±1.74 | 65.00±1.35 | 76.07±1.52 | 65.86±1.05 |
| text-davinci-002 | 76.60±8.77 | 56.50±3.50 | 77.23±9.03 | 76.37±8.61 | 76.73±8.78 | 74.77±7.39 | 76.93±8.66 | 67.77±8.14 | 76.87±8.70 | 64.87±11.20 |
| text-davinci-003 | 77.85±1.24 | 57.53±0.46 | 78.41±1.15 | 78.61±1.19 | 77.93±1.18 | 74.29±1.19 | 77.86±1.15 | 69.04±1.49 | 77.86±1.19 | 66.23±1.75 |
| gpt-3.5-turbo | 63.63±0.36 | 42.54±1.82 | 65.51±0.35 | 63.83±1.06 | 61.75±0.75 | 54.12±1.38 | 61.44±1.31 | 49.58±0.38 | 61.31±0.72 | 36.81±0.86 |

Table 6: Performance and robustness test results (micro-F1) of GPT series models in zero-shot and few-shot scenarios on **SQuAD2.0** dataset.

| Model | AddSentDiverse # 5129 samples | | ModifyPos # 5053 samples | | PerturbAnswer # 5522 samples | | PerturbQuestion-BackTranslation # 11492 samples | | PerturbQuestion-MLM # 11491 samples | |
|---|---|---|---|---|---|---|---|---|---|---|
| | ori | trans | ori | trans | ori | trans | ori | trans | ori | trans |
| *0-shot* | | | | | | | | | | |
| code-davinci-002 | **83.94±1.91** | 59.51±2.70 | **81.36±1.86** | 79.55±2.46 | **83.61±2.05** | 80.18±1.24 | 78.90±5.21 | 75.93±5.69 | 78.90±5.21 | 51.54±1.74 |
| text-davinci-001 | 65.36±13.26 | 49.07±15.05 | 65.22±12.79 | 65.55±12.61 | 64.60±13.07 | 59.56±11.98 | 62.61±12.37 | 52.86±8.30 | 62.67±12.33 | 50.51±10.25 |
| text-davinci-002 | 69.31±12.21 | 48.92±14.03 | 69.05±12.28 | 68.67±12.56 | 68.69±12.14 | 63.83±11.56 | 67.36±11.41 | 56.78±9.97 | 67.50±11.23 | **55.52±13.31** |
| text-davinci-003 | 65.42±9.22 | 54.21±9.96 | 66.31±9.45 | 66.16±8.65 | 65.87±9.20 | 65.19±7.96 | 65.96±9.17 | 58.09±7.34 | 66.00±9.11 | 54.87±8.62 |
| gpt-3.5-turbo | 53.88±8.87 | 36.66±4.97 | 55.58±9.04 | 55.14±8.36 | 54.52±8.45 | 47.64±6.84 | 55.67±8.56 | 45.74±6.85 | 55.48±8.59 | 35.86±5.30 |
| *1-shot* | | | | | | | | | | |
| code-davinci-002 | 90.89±0.51 | 67.46±2.57 | 89.48±0.21 | 88.09±1.07 | 90.56±0.19 | 90.43±2.29 | 88.95±0.00 | 83.46±0.78 | 88.95±0.00 | 77.57±4.08 |
| text-davinci-001 | 80.02±1.33 | 63.38±4.22 | 79.96±1.75 | 77.94±1.67 | 79.15±1.28 | 74.07±1.47 | 76.99±1.03 | 63.77±0.97 | 77.00±1.18 | 62.93±1.50 |
| text-davinci-002 | 61.16±17.04 | 48.01±12.27 | 61.14±16.83 | 59.61±16.83 | 61.21±16.64 | 53.20±15.28 | 63.73±14.55 | 55.95±13.82 | 63.67±14.47 | 53.01±14.75 |
| text-davinci-003 | 87.50±1.26 | **69.80±2.07** | 88.01±1.26 | 87.80±1.14 | 87.34±1.16 | 84.79±1.07 | 87.24±1.30 | 79.16±1.14 | 87.36±1.24 | 77.04±1.67 |
| gpt-3.5-turbo | 78.81±2.35 | 57.64±3.23 | 79.90±2.17 | 79.30±2.13 | 78.28±2.53 | 70.80±2.18 | 78.79±2.56 | 66.04±1.54 | 79.03±2.35 | 52.84±1.75 |
| *3-shot* | | | | | | | | | | |
| code-davinci-002 | **92.90±1.11** | 61.40±2.60 | **92.87±1.85** | 90.91±1.31 | **92.56±0.59** | 92.87±1.12 | 91.54±0.44 | 85.92±1.91 | 91.54±0.44 | 77.88±0.43 |
| text-davinci-001 | 81.97±1.67 | 60.88±1.33 | 81.53±1.65 | 80.49±2.60 | 80.63±1.58 | 75.58±0.85 | 77.99±1.47 | 66.04±2.09 | 77.94±1.52 | 64.38±0.83 |
| text-davinci-002 | 84.44±5.89 | 57.83±2.05 | 83.64±6.19 | 81.63±7.04 | 83.27±5.90 | 77.73±7.09 | 84.83±4.34 | 73.77±5.07 | 84.67±4.69 | 71.42±7.39 |
| text-davinci-003 | 88.88±0.38 | 67.08±0.90 | 89.29±0.37 | 89.41±0.47 | 88.83±0.43 | 86.36±0.42 | 88.88±0.36 | 80.89±0.51 | 88.87±0.46 | **78.87±1.03** |
| gpt-3.5-turbo | 79.68±0.68 | 57.46±0.37 | 80.62±0.56 | 80.45±0.56 | 78.97±0.26 | 71.83±0.72 | 80.03±0.21 | 66.07±0.65 | 79.87±0.28 | 52.18±0.75 |

Table 7: Performance and robustness test results (Exact Match) of GPT series models in zero-shot and few-shot scenarios on **SQuAD2.0** dataset.

| Model | AddSentDiverse # 5129 samples | | ModifyPos # 5053 samples | | PerturbAnswer # 5522 samples | | PerturbQuestion-BackTranslation # 11492 samples | | PerturbQuestion-MLM # 11491 samples | |
|---|---|---|---|---|---|---|---|---|---|---|
| | ori | trans | ori | trans | ori | trans | ori | trans | ori | trans |
| *0-shot* | | | | | | | | | | |
| code-davinci-002 | **72.67±2.52** | **50.33±2.52** | **68.67±2.08** | **65.67±2.89** | **72.33±3.06** | 69.00±1.00 | 66.67±3.55 | 60.47±4.66 | 66.67±3.55 | **41.86±2.33** |
| text-davinci-001 | 45.60±18.09 | 31.87±18.18 | 45.37±17.68 | 46.07±17.56 | 44.67±17.72 | 39.73±15.84 | 43.21±16.90 | 35.46±12.67 | 43.21±16.90 | 32.92±12.81 |
| text-davinci-002 | 50.73±17.72 | 33.03±17.64 | 50.13±17.98 | 49.70±18.20 | 49.83±17.31 | 44.47±16.12 | 49.04±15.81 | 40.26±14.91 | 49.11±15.68 | 38.00±16.24 |
| text-davinci-003 | 40.55±14.61 | 31.78±15.33 | 41.63±15.09 | 41.54±14.10 | 41.34±14.56 | 41.07±12.80 | 41.50±14.42 | 34.33±11.71 | 41.43±14.52 | 31.76±13.05 |
| gpt-3.5-turbo | 25.90±12.45 | 10.97±6.61 | 27.58±12.99 | 26.54±11.90 | 26.08±11.85 | 20.02±9.10 | 27.50±11.95 | 19.72±9.87 | 27.59±12.19 | 14.63±6.92 |
| *1-shot* | | | | | | | | | | |
| code-davinci-002 | 82.67±1.15 | **60.25±2.63** | 82.00±2.71 | 79.75±2.36 | 83.25±1.89 | 84.00±3.65 | 77.32±1.16 | 71.51±2.23 | 77.32±1.16 | **68.61±3.00** |
| text-davinci-001 | 65.43±2.02 | 50.03±2.97 | 64.93±2.26 | 63.67±2.31 | 63.93±2.05 | 58.83±1.81 | 61.66±1.37 | 49.66±1.73 | 61.59±1.66 | 48.15±1.03 |
| text-davinci-002 | 49.87±16.57 | 38.13±12.02 | 49.67±16.31 | 48.27±16.34 | 49.70±16.10 | 41.50±13.64 | 51.51±14.46 | 43.62±13.46 | 51.51±14.33 | 40.40±14.94 |
| text-davinci-003 | 73.00±3.19 | 56.06±3.08 | 73.95±3.16 | 74.08±2.88 | 72.73±2.98 | 69.40±2.60 | 72.83±3.42 | 63.54±2.93 | 73.04±3.25 | 60.72±3.07 |
| gpt-3.5-turbo | 59.19±3.63 | 39.75±4.73 | 60.78±3.79 | 60.31±3.71 | 57.99±3.96 | 51.14±3.80 | 58.91±4.19 | 46.43±2.89 | 59.13±3.71 | 35.76±3.18 |
| *3-shot* | | | | | | | | | | |
| code-davinci-002 | **85.67±1.15** | 56.33±2.08 | **84.67±2.31** | 82.33±1.15 | **85.33±0.58** | **87.33±0.58** | 79.85±1.35 | 74.42±2.33 | 79.85±1.35 | 68.22±1.35 |
| text-davinci-001 | 69.75±2.42 | 49.47±0.67 | 68.91±2.49 | 68.27±3.15 | 67.87±2.22 | 62.97±1.45 | 65.16±1.38 | 54.44±1.99 | 65.09±1.73 | 51.45±1.10 |
| text-davinci-002 | 74.87±6.59 | 49.60±2.60 | 73.57±6.83 | 71.57±7.80 | 73.03±6.54 | 66.80±7.30 | 74.07±5.41 | 62.83±6.49 | 73.87±5.64 | 59.19±7.75 |
| text-davinci-003 | 76.20±1.29 | 55.70±0.47 | 76.96±1.17 | 77.57±1.43 | 76.29±1.25 | 72.95±1.24 | 76.42±1.17 | 67.01±1.46 | 76.43±1.33 | 64.19±1.94 |
| gpt-3.5-turbo | 61.81±1.07 | 41.32±1.50 | 63.08±0.86 | 62.72±1.06 | 59.99±0.93 | 53.44±1.47 | 61.61±0.74 | 48.00±0.42 | 61.40±0.75 | 36.40±1.14 |

For the MRC task, we selecte two datasets, SQuAD1.1 and SQuAD2.0, and use two evaluation metrics, F1 and EM, for each dataset. We analyze the performance and robustness of the GPT model in both zero-shot and few-shot scenarios below. More details are shown in Table 4 to 7

**In the zero-shot scenario, code-davinci-002 achieves the best performance.** It is obvious that code-davinci-002 shows much better performance than the other four models whether using the F1 or EM evaluation metric. It is worth mentioning that although gpt-3.5-turbo has poor results in both evaluation metrics in the zero-shot scenario, it does not necessarily mean that the model performs poorly in the MRC task. The reason is that gpt-3.5-turbo is a chat-oriented model, which tends to generate more complete sentences. Although these sentences often contain the correct answer, the limitations of automatic evaluation metrics result in the model scoring lower in the zero-shot scenario.

**In the few-shot scenario, code-davinci-002's performance is still impressive, especially in the three-shot scenario where it outperforms the other four models by ten to twenty points.** Meanwhile, with the examples in the prompt, gpt-3.5-turbo is able to give answer words or phrases instead of complete sentences in the zero-shot scenario, resulting in a significant improvement in performance metrics. However, there is still a certain gap compared to code-davinci-002, text-davinci-003, and even text-davinci-002.

**Unfortunately, despite the generational updates of the GPT series models, their robustness in the MRC task has not significantly changed.**

### 4.2.3 Named Entity Recognition

Table 8: Performance and robustness test results (micro-F1) of GPT series models in zero-shot and few-shot scenarios on **ACE2005** dataset.

| Model | ConcatSent # 1312 samples | | CrossCatagory # 1312 samples | | EntTypos # 1405 samples | | OOV # 1312 samples | | SwapLonger # 1312 samples | |
|---|---|---|---|---|---|---|---|---|---|---|
| | ori | trans | ori | trans | ori | trans | ori | trans | ori | trans |
| *0-shot* | | | | | | | | | | |
| code-davinci-002 | 25.08±2.12 | 22.36±1.59 | 24.97±2.64 | **39.86±0.70** | 28.56±1.62 | 25.45±0.91 | 24.98±1.90 | 57.33±4.44 | 25.64±2.75 | 66.83±4.88 |
| text-davinci-001 | 12.71±5.81 | 12.75±2.83 | 12.68±5.77 | 10.50±3.93 | 15.93±4.26 | 14.54±2.22 | 12.86±5.85 | 17.18±5.71 | 12.81±5.87 | 20.93±8.03 |
| text-davinci-002 | 35.50±1.64 | **37.95±11.49** | 32.65±3.89 | 30.58±1.64 | **37.13±0.86** | **35.22±1.36** | 35.55±1.57 | 59.20±1.87 | 32.60±3.97 | 65.36±1.23 |
| text-davinci-003 | **36.93±1.86** | 29.43±1.99 | **37.03±1.69** | 36.33±1.17 | 31.31±2.35 | 26.42±2.31 | **36.97±1.91** | 69.86±2.92 | **36.90±1.85** | 75.39±2.20 |
| gpt-3.5-turbo | 32.68±3.24 | 29.52±0.18 | 34.68±0.22 | 36.38±0.65 | 30.97±1.57 | 27.10±1.66 | 34.49±0.28 | **70.07±1.17** | 34.40±0.42 | **78.41±1.21** |
| *1-shot* | | | | | | | | | | |
| code-davinci-002 | **53.69±3.42** | **40.51±5.03** | **53.60±3.14** | **50.69±1.26** | **44.98±2.24** | **40.93±4.35** | **53.78±3.49** | **79.73±1.73** | **53.55±3.30** | **86.99±0.42** |
| text-davinci-001 | 23.46±3.10 | 22.58±4.43 | 23.50±3.25 | 17.31±0.43 | 25.61±2.04 | 21.45±2.66 | 23.49±3.14 | 31.72±0.39 | 23.49±3.01 | 38.48±0.83 |
| text-davinci-002 | 45.81±0.78 | 39.12±0.69 | 45.71±0.87 | 35.24±0.69 | 40.50±0.34 | 37.27±0.41 | 45.77±0.73 | 64.53±0.10 | 45.54±0.74 | 73.42±0.95 |
| text-davinci-003 | 43.37±1.52 | 34.68±1.61 | 43.48±1.48 | 34.94±0.45 | 37.63±1.28 | 30.82±1.32 | 43.43±1.41 | 70.35±1.05 | 43.34±1.49 | 77.42±0.36 |
| gpt-3.5-turbo | 39.39±2.67 | 34.15±0.86 | 39.88±2.36 | 37.78±1.09 | 38.63±0.57 | 33.12±1.21 | 39.29±2.83 | 71.21±2.02 | 39.35±2.97 | 77.04±1.91 |
| *3-shot* | | | | | | | | | | |
| code-davinci-002 | **67.48±1.47** | **65.21±0.06** | **65.20±3.31** | 42.69±0.67 | **52.27±1.83** | **49.94±0.85** | **67.31±0.30** | 61.33±2.68 | **65.32±1.58** | 47.94±16.00 |
| text-davinci-001 | 26.56±1.93 | 26.46±1.27 | 26.48±2.11 | 16.50±0.43 | 25.06±1.56 | 23.81±1.18 | 26.57±2.01 | 30.24±1.99 | 26.75±2.06 | 28.47±2.60 |
| text-davinci-002 | 35.68±3.97 | 32.92±4.99 | 35.82±3.89 | 20.12±4.07 | 30.70±2.24 | 35.52±2.08 | 35.85±4.15 | 43.46±2.00 | 35.90±4.11 | 42.87±4.30 |
| text-davinci-003 | 54.11±0.60 | 47.30±1.01 | 53.99±0.59 | **39.21±0.26** | 45.51±0.44 | 43.10±0.24 | 54.29±0.51 | **71.46±0.22** | 54.17±0.77 | **77.44±0.37** |
| gpt-3.5-turbo | 48.46±0.93 | 40.43±1.51 | 48.43±1.14 | 38.93±0.16 | 43.32±0.71 | 38.99±1.97 | 48.40±0.77 | 70.16±0.49 | 48.40±0.82 | 77.76±0.39 |

Table 9: Performance and robustness test results (micro-F1) of GPT series models in zero-shot and few-shot scenarios on **CoNLL2003** dataset.

| Model | ConcatSent # 3453 samples | | CrossCatagory # 3453 samples | | EntTypos # 2676 samples | | OOV # 3453 samples | | SwapLonger # 3453 samples | |
|---|---|---|---|---|---|---|---|---|---|---|
| | ori | trans | ori | trans | ori | trans | ori | trans | ori | trans |
| *0-shot* | | | | | | | | | | |
| code-davinci-002 | **69.04±3.27** | 58.28±15.34 | **69.17±3.97** | 29.00±11.44 | 41.94±22.37 | **57.92±8.76** | **56.53±24.73** | **56.65±13.45** | **60.85±11.06** | 48.65±1.28 |
| text-davinci-001 | 18.92±1.12 | 18.48±1.64 | 18.84±1.48 | 11.94±0.45 | 22.00±0.97 | 18.61±1.24 | 18.72±0.95 | 13.83±1.35 | 19.12±1.17 | 16.36±1.07 |
| text-davinci-002 | 54.69±1.28 | 58.04±0.93 | 54.91±1.41 | **29.19±0.52** | **57.04±0.75** | 49.04±0.66 | 54.79±1.37 | 55.25±0.51 | 54.79±1.28 | **52.14±0.61** |
| text-davinci-003 | 50.46±2.10 | 52.97±1.27 | 50.47±2.15 | 26.39±1.50 | 55.93±1.81 | 43.43±1.31 | 50.42±2.21 | 47.13±1.55 | 50.43±2.14 | 43.29±1.32 |
| gpt-3.5-turbo | 43.39±2.72 | 49.45±2.07 | 43.34±2.68 | 22.81±2.09 | 49.49±3.24 | 39.40±2.36 | 43.52±2.81 | 40.40±3.55 | 43.30±2.65 | 34.41±2.65 |
| *1-shot* | | | | | | | | | | |
| code-davinci-002 | **72.83±5.37** | **69.33±12.28** | **73.01±5.83** | **43.94±5.68** | **71.42±4.60** | **64.34±4.00** | **71.82±7.03** | 48.93±11.49 | **72.95±5.46** | 57.98±1.29 |
| text-davinci-001 | 28.66±1.42 | 26.06±2.43 | 28.07±1.50 | 19.39±0.85 | 30.85±1.38 | 27.80±1.69 | 28.75±1.44 | 28.04±2.16 | 28.44±1.17 | 33.17±2.45 |
| text-davinci-002 | 56.70±1.50 | 55.76±2.09 | 56.73±1.10 | 35.40±1.22 | 58.70±1.71 | 51.53±2.57 | 56.69±1.44 | **58.46±1.97** | 56.82±1.36 | **59.20±0.82** |
| text-davinci-003 | 52.40±1.24 | 53.96±1.98 | 52.35±1.29 | 27.11±1.72 | 57.54±1.14 | 47.84±1.11 | 52.39±1.11 | 47.95±1.25 | 52.36±1.13 | 46.13±1.77 |
| gpt-3.5-turbo | 50.56±1.82 | 52.88±1.93 | 50.47±1.79 | 27.59±1.26 | 55.04±2.07 | 46.35±2.87 | 50.49±1.74 | 47.54±1.14 | 50.57±1.65 | 47.66±0.28 |
| *3-shot* | | | | | | | | | | |
| code-davinci-002 | 54.54±2.07 | 48.96±4.85 | 54.62±1.69 | 31.30±1.69 | 56.32±1.84 | 50.41±0.13 | 54.83±1.63 | 44.99±1.95 | 54.67±1.96 | 39.39±4.36 |
| text-davinci-001 | 35.69±5.59 | 30.10±5.72 | 35.53±5.95 | 19.32±6.15 | 37.50±4.41 | 29.81±4.40 | 35.54±4.58 | 28.60±4.87 | 35.71±4.46 | 39.36±4.14 |
| text-davinci-002 | **61.64±0.87** | **59.15±0.84** | **61.52±0.94** | **33.21±0.15** | **63.20±0.93** | **54.58±1.83** | **61.50±0.84** | **64.52±0.27** | **61.52±0.92** | **58.78±0.47** |
| text-davinci-003 | 57.73±1.50 | 56.80±1.58 | 57.62±1.58 | 30.93±0.38 | 61.13±1.55 | 49.65±1.40 | 57.70±1.56 | 56.21±1.86 | 57.70±1.60 | 52.34±0.32 |
| gpt-3.5-turbo | 57.74±2.20 | 57.70±1.45 | 57.73±2.27 | 29.99±1.09 | 60.08±1.85 | 49.94±2.59 | 57.51±2.14 | 53.45±1.73 | 57.72±1.79 | 51.57±0.86 |

Table 10: Performance and robustness test results (micro-F1) of GPT series models in zero-shot and few-shot scenarios on **Ontonotesv5** dataset.

| Model | ConcatSent # 4019 samples | | CrossCatagory # 4019 samples | | EntTypos # 4492 samples | | OOV # 4019 samples | | SwapLonger # 4019 samples | |
|---|---|---|---|---|---|---|---|---|---|---|
| | ori | trans | ori | trans | ori | trans | ori | trans | ori | trans |
| *0-shot* | | | | | | | | | | |
| code-davinci-002 | 0.00±0.00 | 0.00±0.00 | 0.00±0.00 | 2.17±0.48 | 16.96±1.07 | 15.26±0.89 | 0.00±0.00 | 7.45±0.61 | 0.00±0.00 | 3.44±1.51 |
| text-davinci-001 | 0.17±0.12 | 0.21±0.04 | 0.17±0.12 | 0.22±0.14 | 15.14±1.98 | 12.18±0.79 | 0.17±0.12 | 0.78±0.07 | 0.15±0.12 | 0.37±0.07 |
| text-davinci-002 | 1.94±0.25 | 2.17±0.45 | 2.33±0.64 | 1.05±0.33 | 29.78±0.82 | 25.82±0.67 | 1.98±0.20 | 3.95±0.91 | 1.97±0.31 | 2.87±0.45 |
| text-davinci-003 | 7.19±0.58 | 7.76±0.73 | 7.11±0.56 | 3.30±0.25 | 31.10±2.69 | 25.32±2.22 | 7.13±0.50 | 13.05±0.19 | 7.16±0.46 | 12.05±0.98 |
| gpt-3.5-turbo | **13.68±0.69** | **16.27±1.56** | **13.75±0.71** | **7.47±0.14** | **34.15±0.20** | **27.73±0.76** | **13.60±0.53** | **21.91±0.80** | **13.53±0.55** | **20.05±0.65** |
| *1-shot* | | | | | | | | | | |
| code-davinci-002 | 14.06±1.45 | 10.46±9.06 | 13.81±1.86 | 12.01±0.43 | 26.80±1.74 | 23.21±2.35 | 14.06±1.45 | 51.24±3.62 | 14.06±1.45 | 39.17±1.44 |
| text-davinci-001 | 0.87±0.24 | 1.24±0.24 | 0.80±0.23 | 0.50±0.15 | 10.57±0.59 | 8.26±0.92 | 0.83±0.25 | 1.47±0.09 | 0.81±0.22 | 1.20±0.23 |
| text-davinci-002 | 4.22±0.33 | 5.06±0.16 | 4.09±0.11 | 2.44±0.39 | 29.29±3.65 | 27.95±2.47 | 4.07±0.29 | 8.93±0.60 | 4.22±0.38 | 7.01±0.64 |
| text-davinci-003 | 13.54±1.29 | 15.04±1.53 | 13.53±1.37 | 6.32±0.41 | **37.40±1.49** | **31.08±1.44** | 13.53±1.35 | 18.76±1.26 | 13.57±1.32 | 17.75±1.38 |
| gpt-3.5-turbo | 17.22±0.87 | **19.99±0.65** | 17.19±0.86 | 8.74±0.98 | 37.05±0.53 | 30.00±1.12 | 17.23±0.82 | 25.24±1.25 | 17.10±1.02 | 24.22±1.18 |
| *3-shot* | | | | | | | | | | |
| code-davinci-002 | 15.38±0.00 | 16.24±0.74 | 15.38±0.00 | **14.29±0.00** | 29.23±0.61 | 26.91±2.61 | 15.38±0.00 | **71.43±0.00** | 15.38±0.00 | **40.00±0.00** |
| text-davinci-001 | 2.25±1.61 | 2.23±1.38 | 2.25±1.61 | 2.53±1.88 | 16.70±3.56 | 13.60±3.21 | 2.31±1.71 | 6.16±4.56 | 2.23±1.58 | 4.55±3.28 |
| text-davinci-002 | 8.00±3.11 | 9.78±1.14 | 8.59±2.19 | 5.43±1.10 | 33.50±7.37 | 26.68±5.88 | 8.57±2.17 | 17.96±2.36 | 8.46±2.77 | 14.90±2.97 |
| text-davinci-003 | 16.63±1.85 | 19.07±1.02 | 16.61±1.77 | 7.85±1.04 | 36.85±1.87 | 31.04±2.01 | 16.64±1.76 | 24.47±2.11 | 16.68±1.85 | 23.13±1.99 |
| gpt-3.5-turbo | **18.22±0.56** | 19.97±0.56 | **18.26±0.65** | 9.63±0.17 | **37.38±0.38** | **30.40±0.84** | **18.24±0.61** | 26.32±0.48 | **18.22±0.72** | 25.62±0.63 |

Table 11: Performance (micro-F1) of GPT series models in zero-shot and few-shot scenarios on **HONOR**, **MSRANER**, **OntoNote4NER** dataset.

| Model | HONOR<br># 1120 samples | MSRANER<br># 4365 samples | OntoNote4NER<br># 4346 samples |
|---|---|---|---|
| *0-shot* | | | |
| code-davinci-002 | 43.12±2.02 | 10.59±0.43 | 12.74±4.12 |
| text-davinci-001 | - | - | - |
| text-davinci-002 | 46.60±8.34 | 15.13±1.83 | - |
| text-davinci-003 | 47.62±2.00 | **23.02±4.67** | **30.66±1.19** |
| gpt-3.5-turbo | **50.85±0.64** | 20.19±6.45 | 11.12±1.36 |
| *1-shot* | | | |
| code-davinci-002 | 54.89±0.60 | 56.78±14.81 | 33.96±11.91 |
| text-davinci-001 | 32.42±1.05 | 25.16±2.18 | - |
| text-davinci-002 | 45.14±11.55 | 35.63±2.14 | 35.69±4.34 |
| text-davinci-003 | 49.04±0.89 | 46.89±0.81 | 49.72±0.77 |
| gpt-3.5-turbo | 47.87±2.41 | 11.39±5.73 | 36.48±10.97 |
| *3-shot* | | | |
| code-davinci-002 | **60.95±0.36** | 53.23±10.25 | 34.58±0.72 |
| text-davinci-001 | 39.09±0.40 | 42.43±0.68 | 36.38±0.23 |
| text-davinci-002 | 51.02±13.63 | **58.24±1.88** | 34.27±11.17 |
| text-davinci-003 | 54.01±2.22 | 57.14±0.36 | **50.98±0.94** |
| gpt-3.5-turbo | 55.72±0.33 | 52.53±3.13 | 35.77±2.38 |

We analyze the performance of six different NER datasets on various models and find that each model has its own unique characteristics. Our analysis includes two scenarios: zero-shot and few-shot. For details of the results, please refer to Table 8 ∼ Table 11.

**In the zero-shot scenario, the text-davinci-003 and gpt-3.5-turbo models consistently performed the best.** The ACE2005 dataset achieves its best performance on text-davinci-002 or text-davinci-003, with gpt-3.5-turbo performing similarly well. Code-davinci-002 performs best on the CoNLL2003 dataset, while the OntonotesV5 dataset achieves its highest performance on gpt-3.5-turbo. For the HONOR dataset, gpt-3.5-turbo has the best performance, whereas the MSRANER and Ontonote4NER datasets perform best on text-davinci-003.

**In the few-shot scenario, each model performs differently on different datasets.** The three-shot code-davinci-002 achieves the best performance on the ACE2005 dataset, indicating that increasing the number of examples greatly improve the model's performance on this dataset. On the CoNLL2003 dataset, the 1-shot code-davinci-002 model has the best performance. Interestingly, increasing the number of examples on this dataset actually decreases the performance of code-davinci-002 (three-shot performance is lower than one-shot). The Ontonotes v5 dataset achieves its best performance on text-davinci-003, gpt-3.5-turbo, code-davinci-002, and gpt-3.5-turbo. In the HONOR dataset, MSRANER dataset, and Ontonote4NER, the best performance appear on code-davinci-002, text-davinci-002, and text-davinci-003, respectively.

**Despite the intergenerational updates of the GPT series models, we found that there were no significant changes in their robustness.**

### 4.2.4 Natural Language Inference

Table 12: Performance and robustness test results (accuracy) of GPT series models in zero-shot and few-shot scenarios on **MNLI-m** dataset.

| Model | AddSent # 9815 samples | | NumWord # 745 samples | | SwapAnt # 199 samples | |
|---|---|---|---|---|---|---|
| | ori | trans | ori | trans | ori | trans |
| *0-shot* | | | | | | |
| code-davinci-002 | 48.38±3.06 | 37.13±2.43 | 45.08±3.74 | 25.81±19.73 | 76.33±19.40 | 51.33±32.58 |
| text-davinci-001 | 42.20±4.15 | 36.66±2.06 | 38.91±3.03 | 25.76±8.35 | 42.41±23.67 | 28.89±23.00 |
| text-davinci-002 | 52.62±6.64 | 36.27±1.61 | 52.72±5.83 | 29.21±14.81 | 52.27±15.07 | 50.95±19.87 |
| text-davinci-003 | 64.26±0.53 | 34.04±1.85 | 68.12±1.42 | **51.62±13.17** | 70.11±1.57 | **78.96±6.20** |
| gpt-3.5-turbo | **68.96±2.93** | **58.83±4.30** | **70.07±2.46** | 22.37±6.82 | **80.07±9.73** | 44.06±9.43 |
| *1-shot* | | | | | | |
| code-davinci-002 | 72.67±3.06 | 39.67±4.04 | 70.92±3.51 | **54.33±15.95** | 76.67±9.50 | **81.00±5.29** |
| text-davinci-001 | 39.90±8.01 | 39.40±7.63 | 40.44±7.19 | 1.39±2.40 | **98.16±3.19** | 4.36±6.28 |
| text-davinci-002 | 68.60±6.20 | 37.20±1.10 | 69.44±9.26 | 45.37±14.51 | 90.29±1.05 | **86.26±10.18** |
| text-davinci-003 | **73.14±4.80** | 41.07±5.05 | **74.43±4.90** | 43.80±3.61 | 93.35±1.66 | 72.87±5.45 |
| gpt-3.5-turbo | 71.69±0.31 | **57.77±3.91** | 71.32±1.30 | 43.04±6.98 | 92.46±1.81 | 74.70±5.13 |
| *3-shot* | | | | | | |
| code-davinci-002 | **74.99±6.10** | 41.67±3.51 | 67.81±7.67 | **55.00±7.81** | 75.74±7.61 | 87.67±9.29 |
| text-davinci-001 | 48.80±3.64 | 43.87±3.91 | 50.16±6.62 | 4.34±5.34 | 91.79±7.33 | 28.64±16.34 |
| text-davinci-002 | 70.30±5.47 | 36.87±2.12 | 71.18±6.00 | 44.61±11.68 | 97.49±1.51 | **88.27±12.51** |
| text-davinci-003 | 72.07±5.69 | 41.02±3.14 | 71.59±6.01 | 46.49±4.12 | **98.66±0.58** | 82.41±5.10 |
| gpt-3.5-turbo | 68.98±1.53 | 42.69±4.75 | 69.71±1.57 | 46.98±1.91 | 94.64±0.77 | 83.25±2.77 |

Table 13: Performance and robustness test results (accuracy) of GPT series models in zero-shot and few-shot scenarios on **MNLI-mm** dataset.

| Model | AddSent # 9832 samples | | NumWord # 775 samples | | SwapAnt # 255 samples | |
|---|---|---|---|---|---|---|
| | ori | trans | ori | trans | ori | trans |
| *0-shot* | | | | | | |
| code-davinci-002 | 48.49±1.82 | 43.50±5.22 | 50.67±2.89 | 22.67±17.16 | **81.00±16.64** | 55.67±31.07 |
| text-davinci-001 | 44.07±3.54 | 35.72±3.46 | 45.34±6.09 | 28.11±9.66 | 33.42±18.59 | 31.07±18.61 |
| text-davinci-002 | 49.93±6.44 | 36.05±2.27 | 52.93±7.57 | 28.45±14.29 | 51.40±15.02 | 58.39±23.61 |
| text-davinci-003 | 64.56±0.32 | 34.26±2.15 | 67.50±0.25 | **39.81±3.04** | 67.76±0.54 | **77.65±2.71** |
| gpt-3.5-turbo | **69.24±2.21** | **60.37±3.42** | **69.09±1.94** | 19.18±8.02 | 78.30±12.00 | 43.27±9.97 |
| *1-shot* | | | | | | |
| code-davinci-002 | 57.14±14.48 | 43.88±10.34 | 61.00±15.72 | **27.36±17.51** | 79.25±11.63 | 54.88±28.82 |
| text-davinci-001 | 47.54±2.85 | 44.87±2.63 | 45.22±2.84 | 11.10±5.54 | 91.76±8.38 | 36.69±6.32 |
| text-davinci-002 | 51.12±18.88 | 35.20±6.03 | 52.96±20.00 | 16.95±7.03 | 55.30±36.10 | 51.29±18.91 |
| text-davinci-003 | 70.28±5.21 | 37.01±3.34 | 71.39±4.39 | 28.45±1.80 | **93.20±2.94** | **66.80±6.64** |
| gpt-3.5-turbo | **72.96±1.53** | **53.49±3.24** | **72.78±1.40** | 19.53±2.91 | 81.05±4.84 | 53.85±6.61 |
| *3-shot* | | | | | | |
| code-davinci-002 | 68.15±11.92 | 48.93±8.10 | 62.79±7.14 | 47.09±31.37 | 63.33±40.69 | **90.82±3.91** |
| text-davinci-001 | 54.43±3.50 | 49.03±10.00 | 52.16±5.08 | 7.87±10.58 | 89.41±9.90 | 32.29±23.66 |
| text-davinci-002 | 61.81±11.09 | 34.79±0.70 | 61.40±14.05 | **49.71±20.48** | 70.67±19.95 | 83.12±26.82 |
| text-davinci-003 | **73.66±3.44** | 37.92±1.28 | **74.99±2.36** | 41.59±4.71 | **96.34±1.20** | 89.61±2.90 |
| gpt-3.5-turbo | 73.20±1.55 | 42.94±5.07 | 74.88±1.56 | 46.80±5.25 | 90.20±3.14 | 82.22±4.65 |

Table 14: Performance and robustness test results (accuracy) of GPT series models in zero-shot and few-shot scenarios on **SNLI** dataset.

| Model | AddSent # 10000 samples | | NumWord # 108 samples | | SwapAnt # 523 samples | |
|---|---|---|---|---|---|---|
| | ori | trans | ori | trans | ori | trans |
| *0-shot* | | | | | | |
| code-davinci-002 | 56.67±4.51 | 38.33±5.86 | 52.00±4.36 | 55.33±26.76 | 74.67±17.47 | 60.00±25.12 |
| text-davinci-001 | 38.61±7.26 | 33.00±3.80 | 48.30±26.79 | 46.42±30.83 | 24.75±15.46 | **63.38±37.26** |
| text-davinci-002 | 47.40±8.24 | 35.45±2.04 | 45.98±11.81 | 30.84±25.56 | 58.76±41.98 | 27.89±20.28 |
| text-davinci-003 | **67.16±3.26** | 34.11±0.03 | 63.89±2.45 | **73.77±22.76** | 81.39±8.06 | 61.44±17.44 |
| gpt-3.5-turbo | 62.57±1.94 | **49.40±2.69** | **68.57±5.08** | 51.23±12.09 | **88.75±5.59** | 42.77±9.83 |
| *1-shot* | | | | | | |
| code-davinci-002 | 72.00±14.73 | 39.67±7.37 | 66.67±5.86 | 52.00±7.81 | 59.33±24.79 | 81.67±4.04 |
| text-davinci-001 | 36.27±0.31 | 34.50±0.61 | 41.97±2.33 | 62.66±11.50 | 2.10±3.48 | 38.24±22.93 |
| text-davinci-002 | **72.20±3.04** | 40.70±3.72 | 71.60±2.98 | 46.91±10.65 | 94.45±4.45 | 60.48±13.19 |
| text-davinci-003 | 71.81±4.12 | 34.63±0.26 | **73.30±1.63** | 54.63±6.68 | **97.77±0.67** | 49.46±8.36 |
| gpt-3.5-turbo | 68.97±1.91 | **51.51±3.13** | 68.82±1.07 | **85.49±3.74** | 93.83±1.55 | **87.19±1.99** |
| *3-shot* | | | | | | |
| code-davinci-002 | **74.67±8.08** | 47.67±4.16 | 69.00±2.65 | 68.33±20.40 | 94.33±3.51 | 49.00±10.44 |
| text-davinci-001 | 44.60±4.45 | 34.10±0.78 | 47.33±6.43 | **87.94±11.21** | 41.27±26.07 | 89.68±6.84 |
| text-davinci-002 | 72.43±3.11 | 37.83±2.73 | 70.68±2.97 | 63.27±15.28 | 94.39±3.75 | 66.35±11.72 |
| text-davinci-003 | 72.18±2.37 | 40.00±2.68 | **75.00±3.34** | 67.29±7.54 | 94.65±2.20 | 45.41±7.14 |
| gpt-3.5-turbo | 68.23±1.99 | 38.31±1.75 | 69.14±2.14 | 84.57±2.67 | **94.70±1.12** | **93.31±1.70** |

We analyze the performance of different GPT series models on the MNLI-m, MMLI-mm, and SNLI NLI datasets, and analyze their performance and robustness in both zero-shot and few-shot scenarios. Overall, the performance of different models on the three NLI datasets shows a similar trend. Please refer to Table 12 to Table 14 for more details.

**In the zero-shot scenario, gpt-3.5-turbo performes the best most of the time, followed by text-davinci-003.** Meanwhile, Code-davinci-002 and text-davinci-002 also performe well on a few datasets, such as the SwapAnt variation of the original MNLI-m dataset, where the performance of code-davinci-002 even exceeds that of gpt-3.5-turbo and text-davinci-003, but this good performance is not stable. However, text-davinci-001 performes poorly in most cases, with a significant gap compared to the other four models.

**In the few-shot scenario, the advantage of gpt-3.5-turbo in performance is no longer as significant as in the zero-shot scenario.** Although text-davinci-001 still has a significant gap compared to the other four models, the performance gap among these five models is significantly reduced compared to the zero-shot scenario, and overall the best performer is text-davinci-003. In addition, on the three NLI datasets, different models generally perform better in the three-shot scenario than in the one-shot scenario, indicating that more prompts can help improve the performance of this series of models.

**Surprisingly, the robustness of gpt-3.5-turbo in NLI tasks is often not as good as earlier models.** For example, in the zero-shot scenario, in the zero-shot scenario, gpt-3.5-turbo shows poor robustness on the NumWord variation of all three datasets, and performs much worse than the other four models.

### 4.2.5 Part-of-speech Tagging

Table 15: Performance and robustness test results (accuracy) of GPT series models in zero-shot and few-shot scenarios on **WSJ** dataset.

(a)

| Model | SwapMultiPOSJJ # 3963 samples | | SwapMultiPOSNN # 4952 samples | | SwapMultiPOSRB # 2874 samples | |
|---|---|---|---|---|---|---|
| | ori | trans | ori | trans | ori | trans |
| *0-shot* | | | | | | |
| code-davinci-002 | 43.15±16.40 | 46.25±16.37 | 45.59±15.77 | 47.95±14.79 | 41.63±18.46 | 41.40±13.71 |
| text-davinci-001 | - | - | - | - | - | - |
| text-davinci-002 | 71.48±0.56 | 70.78±0.69 | 71.13±0.67 | 70.06±0.76 | 70.62±0.52 | 69.14±0.77 |
| text-davinci-003 | **75.67±2.65** | **74.92±2.79** | **75.47±2.63** | **74.26±2.72** | **74.81±2.60** | **72.80±2.81** |
| gpt-3.5-turbo | - | - | - | - | - | - |
| *1-shot* | | | | | | |
| code-davinci-002 | 77.70±0.74 | 76.84±0.84 | 77.37±0.78 | 76.73±0.54 | 77.81±0.37 | 74.91±0.78 |
| text-davinci-001 | - | - | - | - | - | - |
| text-davinci-002 | 68.75±0.71 | 67.68±0.55 | 68.29±0.50 | 67.20±0.38 | 68.06±0.51 | 66.26±0.37 |
| text-davinci-003 | 71.42±0.69 | 70.71±0.67 | 71.35±0.64 | 70.46±0.71 | 71.40±0.55 | 69.87±0.54 |
| gpt-3.5-turbo | - | - | - | - | - | - |
| *3-shot* | | | | | | |
| code-davinci-002 | **85.99±0.78** | **85.47±0.61** | **85.34±0.68** | **84.63±0.34** | **85.45±0.77** | **83.09±0.68** |
| text-davinci-001 | - | - | - | - | - | - |
| text-davinci-002 | 79.64±1.00 | 79.28±1.11 | 79.53±1.06 | 78.85±0.95 | 79.74±1.00 | 77.99±0.66 |
| text-davinci-003 | 84.09±0.21 | 83.57±0.17 | 83.91±0.20 | 83.07±0.16 | 83.55±0.29 | 81.86±0.33 |
| gpt-3.5-turbo | 77.59±2.44 | 76.95±2.33 | 77.70±2.35 | 76.55±2.37 | 77.31±2.34 | 75.85±2.08 |

(b)

| Model | SwapMultiPOSVB # 2376 samples | | SwapPrefix # 4526 samples | | all # 5461 samples |
|---|---|---|---|---|---|
| | ori | trans | ori | trans | ori |
| *0-shot* | | | | | |
| code-davinci-002 | 44.61±14.66 | 44.95±12.27 | 44.72±15.05 | 47.49±12.97 | 46.53±17.65 |
| text-davinci-001 | - | - | - | - | - |
| text-davinci-002 | 70.73±0.81 | 70.73±0.81 | 71.43±0.77 | 70.66±0.77 | 71.02±0.62 |
| text-davinci-003 | **76.21±2.61** | **76.21±2.61** | **75.51±2.73** | **74.94±2.74** | **75.02±2.59** |
| gpt-3.5-turbo | - | - | - | - | - |
| *1-shot* | | | | | |
| code-davinci-002 | 78.28±0.32 | 78.21±0.45 | 77.58±0.45 | 76.83±0.36 | 77.50±0.50 |
| text-davinci-001 | - | - | - | - | - |
| text-davinci-002 | 68.91±0.47 | 68.91±0.47 | 68.70±0.34 | 68.21±0.50 | 68.13±0.46 |
| text-davinci-003 | 72.88±0.69 | 72.88±0.69 | 71.46±0.63 | 71.08±0.64 | 70.79±0.71 |
| gpt-3.5-turbo | - | - | - | - | - |
| *3-shot* | | | | | |
| code-davinci-002 | **86.58±0.30** | **86.41±0.40** | **85.67±0.60** | **85.40±0.58** | **85.67±0.64** |
| text-davinci-001 | - | - | - | - | - |
| text-davinci-002 | 81.09±0.92 | 81.09±0.92 | 80.10±1.06 | 79.73±1.02 | 79.48±1.11 |
| text-davinci-003 | 84.61±0.27 | 84.61±0.27 | 84.07±0.20 | 83.67±0.19 | 83.69±0.16 |
| gpt-3.5-turbo | 78.15±2.27 | 77.92±2.19 | 77.83±2.18 | 77.28±2.19 | 77.21±2.40 |

Table 16: Performance of GPT series models in zero-shot and few-shot scenarios on **Daily547** (accuracy) and **PKU-SEGPOS** (micro-F1) dataset.

| Model | Daily547<br># 546 samples | PKU-SEGPOS<br># 5204 samples |
|---|---|---|
| *0-shot* | | |
| code-davinci-002 | 47.21±6.74 | 51.03±0.89 |
| text-davinci-001 | - | - |
| text-davinci-002 | 52.96±4.49 | 39.11±6.12 |
| text-davinci-003 | **64.80±0.18** | **65.86±1.18** |
| gpt-3.5-turbo | - | 52.70±4.76 |
| *1-shot* | | |
| code-davinci-002 | 78.84±0.52 | 75.18±1.00 |
| text-davinci-001 | - | 15.09±0.37 |
| text-davinci-002 | 65.25±0.62 | 56.28±1.55 |
| text-davinci-003 | 77.99±1.15 | 76.43±0.45 |
| gpt-3.5-turbo | - | 82.03±1.05 |
| *3-shot* | | |
| code-davinci-002 | **83.53±0.21** | 79.13±0.69 |
| text-davinci-001 | - | 24.30±0.81 |
| text-davinci-002 | 76.03±0.51 | 51.82±2.00 |
| text-davinci-003 | 82.63±0.55 | 76.88±0.36 |
| gpt-3.5-turbo | 70.83±1.62 | **82.58±0.08** |

We evaluate the performance of the GPT series models on three POS datasets. Please refer to Table 15 to Table 16 for detailed results.

**In the zero-shot scenario, text-davinci-003 performs the best, exhibiting superior performance on all three datasets.** From the tables, we can observe that text-davinci-002 and code-davinci-002 closely follow on the WSJ dataset and the Daily547 dataset, while text-davinci-001 and gpt-3.5-turbo fail to produce output in the expected format. On the PKU-SEGPOS dataset, gpt-3.5-turbo produces output in the expected format and comes second only to text-davinci-003 in terms of performance, with code-davinci-002 performing similarly to gpt-3.5-turbo. Text-davinci-001 has the worst performance, still failing to produce output in the expected format.

**In the few-shot scenario, code-davinci-002 or gpt-3.5-turbo achieve the best performance.** Code-davinci-002 in the three-shot scenario achieves the best performance on the WSJ dataset and the Daily547 dataset, while gpt-3.5-turbo in the three-shot scenario shows the best performance on the PKU-SEGPOS dataset. The different linguistic comprehension that exists in this is to be explored.

**All models that can produce the expected answer (i.e., code-davinci-002, text-davinci-002, text-davinci-003) demonstrate strong robustness on the WSJ dataset.**

### 4.2.6 Relation Extraction

Table 17: Performance and robustness test results (micro-F1) of GPT series models in zero-shot and few-shot scenarios on **Tacred** dataset.

(a)

| Model | InsertClause # 14897 samples | | SwapEnt-LowFreq # 15509 samples | | SwapEnt-MultiType # 15509 samples | | SwapEnt-SamEtype # 15509 samples | |
|---|---|---|---|---|---|---|---|---|
| | ori | trans | ori | trans | ori | trans | ori | trans |
| *0-shot* | | | | | | | | |
| code-davinci-002 | 13.33±4.41 | 12.83±3.55 | 13.89±1.92 | 10.55±2.55 | 12.22±4.20 | 11.11±2.55 | 12.78±2.55 | 11.67±6.01 |
| text-davinci-001 | 10.29±0.73 | 9.27±0.75 | 10.05±0.88 | 11.08±0.91 | 10.15±0.73 | 9.88±1.09 | 10.18±0.81 | 10.43±0.35 |
| text-davinci-002 | 12.34±0.30 | 11.50±0.43 | 12.23±0.09 | 9.19±0.69 | 12.18±0.17 | 9.29±0.81 | 12.23±0.09 | 10.70±0.17 |
| text-davinci-003 | **20.37±1.32** | **18.76±0.88** | **20.40±1.33** | **21.77±1.58** | **20.36±1.24** | **20.23±1.02** | **20.36±1.28** | 2.01±1.63 |
| gpt-3.5-turbo | 14.78±0.73 | 14.29±0.28 | 14.82±0.76 | 15.18±0.47 | 14.83±0.69 | 14.16±0.15 | 14.85±0.69 | **15.16±0.48** |
| *1-shot* | | | | | | | | |
| code-davinci-002 | 16.11±0.96 | 12.78±0.96 | 16.11±0.96 | 17.22±0.96 | 16.11±1.92 | 9.45±2.55 | 15.56±0.96 | 12.78±0.96 |
| text-davinci-001 | 10.57±0.61 | 10.16±0.93 | 10.55±0.60 | 11.62±1.55 | 10.66±0.66 | 10.25±2.08 | 10.50±0.56 | 11.68±1.45 |
| text-davinci-002 | 17.25±0.67 | 14.68±0.68 | 17.03±0.61 | 15.14±0.13 | 17.20±0.61 | 12.50±0.10 | 17.20±0.45 | 15.02±0.87 |
| text-davinci-003 | **23.20±1.63** | **22.22±1.21** | **23.24±1.71** | **23.65±1.72** | **23.25±1.68** | **21.91±1.44** | **23.22±1.72** | **24.06±1.70** |
| gpt-3.5-turbo | 14.86±0.16 | 14.29±0.13 | 14.83±0.15 | 13.86±0.19 | 14.86±0.20 | 12.87±0.25 | 14.86±0.18 | 14.07±0.30 |
| *3-shot* | | | | | | | | |
| code-davinci-002 | 19.44±0.96 | 13.89±1.92 | 20.56±0.96 | 14.44±0.96 | 20.56±0.96 | 10.56±0.96 | 19.44±0.96 | 12.22±0.96 |
| text-davinci-001 | 10.10±0.42 | 9.48±0.19 | 10.15±0.33 | 10.48±0.72 | 9.87±0.19 | 7.44±0.19 | 9.99±0.34 | 10.05±0.66 |
| text-davinci-002 | 17.13±1.14 | 14.48±1.36 | 17.14±1.14 | 14.81±0.35 | 16.85±1.48 | 12.21±1.13 | 17.19±1.33 | 16.11±1.53 |
| text-davinci-003 | 22.01±0.11 | 20.89±0.34 | 21.98±0.18 | 21.53±0.29 | 21.94±0.17 | 19.24±0.73 | 21.96±0.20 | 22.14±0.20 |
| gpt-3.5-turbo | 16.16±0.39 | 15.62±0.27 | 16.15±0.45 | 15.09±0.04 | 16.18±0.40 | 13.33±0.30 | 16.18±0.41 | 15.15±0.28 |

(b)

| Model | SwapTriplePos-Age # 28 samples | | SwapTriplePos-Birth # 48 samples | | SwapTriplePos-Employee # 251 samples | |
|---|---|---|---|---|---|---|
| | ori | trans | ori | trans | ori | trans |
| *0-shot* | | | | | | |
| code-davinci-002 | 58.33±11.48 | 67.86±12.88 | 58.33±5.51 | 56.25±5.51 | **42.00±7.55** | **41.33±10.12** |
| text-davinci-001 | **100.00±0.00** | **100.00±0.00** | 33.90±11.69 | 36.64±7.31 | 0.41±0.41 | 1.22±1.23 |
| text-davinci-002 | 89.29±6.19 | 88.10±2.07 | 56.25±8.33 | 52.08±4.17 | 16.47±7.02 | 17.40±8.52 |
| text-davinci-003 | 96.43±0.00 | **100.00±0.00** | **65.28±1.21** | **71.53±1.21** | 6.53±2.36 | 16.37±3.96 |
| gpt-3.5-turbo | 96.43±3.57 | **100.00±0.00** | 56.25±2.08 | 62.50±2.08 | 8.11±2.67 | 7.18±3.81 |
| *1-shot* | | | | | | |
| code-davinci-002 | 89.29±0.00 | 98.81±2.06 | 55.55±6.02 | 59.72±5.24 | 42.67±0.58 | 42.33±3.06 |
| text-davinci-001 | **100.00±0.00** | **100.00±0.00** | 32.05±9.63 | 35.26±10.15 | 1.44±1.46 | 1.78±1.47 |
| text-davinci-002 | 96.61±0.31 | **100.00±0.00** | 62.50±2.08 | 65.97±3.18 | 47.54±6.79 | 52.32±6.68 |
| text-davinci-003 | **100.00±0.00** | **100.00±0.00** | 62.74±5.92 | 68.31±2.83 | 10.88±1.07 | 16.51±2.26 |
| gpt-3.5-turbo | **100.00±0.00** | **100.00±0.00** | 44.44±8.42 | 50.00±11.60 | 8.37±0.40 | 3.99±0.40 |
| *3-shot* | | | | | | |
| code-davinci-002 | 89.28±6.19 | **100.00±0.00** | 71.53±1.21 | 75.69±6.36 | 34.67±7.02 | 36.00±7.94 |
| text-davinci-001 | **100.00±0.00** | **100.00±0.00** | 47.91±9.55 | 48.61±7.89 | 0.80±1.38 | 0.66±1.15 |
| text-davinci-002 | **100.00±0.00** | **100.00±0.00** | 65.97±13.39 | 63.89±7.89 | **60.03±9.42** | **62.02±6.86** |
| text-davinci-003 | **100.00±0.00** | **100.00±0.00** | **75.00±2.08** | **81.25±0.00** | 19.27±3.72 | 25.02±4.99 |
| gpt-3.5-turbo | **100.00±0.00** | **100.00±0.00** | 41.66±7.22 | 52.78±1.21 | 15.80±3.39 | 12.09±1.88 |

We test the performance of GPT series models on the RE task using the Tacred dataset, and the experimental results can be found in Table 17. **In the zero-shot scenario, text-davinci-003 achieves the best performance in most cases, while gpt-3.5-turbo has the second-best overall performance,** and text-davinci-001 has the worst overall performance. In the few-shot scenario, text-davinci-003 in the one-shot setting achieves the best performance in most cases, and there is only a slight improvement in performance in the three-shot setting compared to the one-shot setting. It is worth noting that in the SwapTriplePos-Age deformation with a small sample size, almost all models can achieve a perfect score **Regarding robustness, code-davinci-002 performs poorly on some deformations, while the other models demonstrate good robustness.**

### 4.2.7 Sentiment Classification

Table 18: Performance and robustness test results (accuracy) of GPT series models in zero-shot and few-shot scenarios on **IMDB** dataset.

| Model | AddSum-Movie # 11257 samples | | AddSum-Person # 12230 samples | | DoubleDenial # 22933 samples | | SwapSpecialEnt-Movie # 11257 samples | | SwapSpecialEnt-Person # 12230 samples | |
|---|---|---|---|---|---|---|---|---|---|---|
| | ori | trans | ori | trans | ori | trans | ori | trans | ori | trans |
| *0-shot* | | | | | | | | | | |
| code-davinci-002 | 88.67±7.57 | 85.67±6.81 | 86.67±9.29 | 80.67±8.50 | 88.33±5.69 | 79.33±8.62 | 88.67±7.57 | 87.67±7.77 | 86.67±9.45 | 86.33±11.24 |
| text-davinci-001 | **92.63±0.54** | **91.59±0.86** | **92.34±0.38** | 68.41±18.64 | 93.20±0.44 | 76.52±15.39 | **92.53±0.47** | 83.91±13.67 | **92.43±0.42** | 81.96±9.59 |
| text-davinci-002 | 91.97±1.27 | 91.17±2.66 | 92.00±1.59 | 87.67±1.85 | **93.33±0.96** | **92.57±1.11** | 91.80±1.41 | 90.90±1.91 | 91.97±1.63 | **91.43±1.97** |
| text-davinci-003 | 91.51±1.14 | 91.56±0.80 | 91.59±1.04 | 89.62±0.57 | 92.02±0.77 | 91.13±0.69 | **91.53±1.16** | **91.17±0.77** | 91.60±0.97 | **91.62±0.86** |
| gpt-3.5-turbo | 91.16±0.31 | 90.63±0.17 | 91.29±0.26 | **90.18±0.49** | 91.78±0.22 | 91.06±0.35 | 91.16±0.33 | 89.42±0.44 | 91.27±0.19 | 90.84±0.23 |
| *1-shot* | | | | | | | | | | |
| code-davinci-002 | **94.67±1.15** | 90.67±1.15 | **92.00±1.73** | 77.67±4.16 | 90.00±1.73 | 82.33±1.53 | **94.67±1.15** | **95.33±2.52** | **92.33±1.53** | 90.00±3.00 |
| text-davinci-001 | 91.93±0.06 | 90.57±0.50 | 91.53±0.51 | 86.26±0.78 | **93.13±0.46** | **92.43±0.57** | 91.57±0.15 | 91.57±0.42 | 91.50±0.35 | 91.33±0.15 |
| text-davinci-002 | 89.53±2.20 | 87.10±3.94 | 89.33±2.03 | 80.93±6.54 | 91.67±1.16 | 89.87±1.91 | 89.53±1.78 | 88.50±2.38 | 89.27±2.05 | 88.70±2.75 |
| text-davinci-003 | 91.67±0.61 | **91.40±0.66** | 91.99±0.53 | **89.86±0.83** | 92.24±0.50 | 90.60±0.75 | 91.68±0.62 | 91.22±0.46 | 92.00±0.51 | **91.87±0.57** |
| gpt-3.5-turbo | 87.28±0.88 | 85.28±1.18 | 87.23±0.72 | 83.17±2.21 | 88.65±0.85 | 84.29±1.28 | 87.24±0.90 | 84.07±0.98 | 87.14±0.68 | 85.58±0.81 |
| *3-shot* | | | | | | | | | | |
| code-davinci-002 | 84.67±0.58 | 79.33±0.58 | 88.00±0.00 | 63.00±0.00 | 88.67±0.58 | 86.33±1.53 | 84.67±0.58 | 84.00±1.00 | 88.67±0.58 | 87.00±1.73 |
| text-davinci-001 | 91.13±0.49 | 89.03±0.32 | 91.40±0.82 | 86.08±0.91 | **93.23±0.49** | **92.40±0.26** | 91.00±0.56 | 90.93±0.45 | 91.77±0.58 | **91.83±0.50** |
| text-davinci-002 | 87.30±0.62 | 84.77±1.01 | 88.53±1.08 | 76.57±2.74 | 91.33±0.90 | 90.23±0.40 | 87.63±0.68 | 86.53±1.25 | 88.37±1.11 | 88.17±0.90 |
| text-davinci-003 | 85.73±3.61 | 84.57±3.83 | 86.22±3.45 | 82.88±3.38 | 87.38±3.10 | 85.80±3.79 | 85.69±3.58 | 85.08±3.82 | 86.23±3.44 | 86.06±3.48 |
| gpt-3.5-turbo | 88.55±0.62 | 87.14±0.57 | 88.91±0.63 | 85.53±0.63 | 89.62±0.70 | 87.08±0.81 | 88.55±0.63 | 86.43±0.35 | 88.91±0.53 | 88.25±0.50 |

We analyze the performance of various models on IMDB datasets in two scenarios: zero-shot and few-shot, and present the experimental results in Table 18.

**In the zero-shot scenario, all models perform well with little variation.** GPT series models show significant performance on the SC task. However, code-davinci-002 and text-davinci-001 exhibit lower robustness compared to the other models.

**In the few-shot scenario, most models perform worse than the zero-shot setting.** A notable observation is that the models produce more irrelevant outputs with more examples in prompts. We speculate that the input text may be too long to affect the model's judgment of contextual information, thereby affecting the accuracy of the model's answer. Besides, code-davinci-002 and text-davinci-001 perform better than other models overall. A possible reason is that other models have weakened their in-context learning ability while increasing instruction alignment.

### 4.2.8 Semantic Matching

Table 19: Performance and robustness test results (accuracy) of GPT series models in zero-shot and few-shot scenarios on **MRPC** dataset.

| Model | NumWord<br># 402 samples | | SwapAnt<br># 158 samples | | all<br># 1724 samples |
|---|---|---|---|---|---|
| | ori | trans | ori | trans | ori |
| *0-shot* | | | | | |
| code-davinci-002 | 0.00±0.00 | 4.67±8.08 | 26.00±45.03 | 8.00±13.86 | 70.00±3.07 |
| text-davinci-001 | 17.58±21.19 | 17.08±17.04 | 22.79±27.59 | 13.71±11.91 | 21.60±19.25 |
| text-davinci-002 | 68.41±6.24 | 66.67±35.79 | **95.57±5.18** | 36.29±18.66 | 72.73±2.55 |
| text-davinci-003 | **74.63±1.97** | **94.44±3.90** | 75.11±8.26 | 54.22±10.41 | 70.17±4.51 |
| gpt-3.5-turbo | 71.57±1.80 | 93.76±2.38 | 89.81±3.37 | **76.22±7.02** | **73.58±0.33** |
| *1-shot* | | | | | |
| code-davinci-002 | 69.00±5.29 | 97.33±3.06 | **89.67±5.51** | 80.33±10.60 | 76.13±3.63 |
| text-davinci-001 | 65.84±3.45 | 78.44±8.76 | 89.66±2.86 | 49.16±7.63 | 70.40±1.87 |
| text-davinci-002 | 72.31±7.04 | 98.59±1.65 | 64.14±14.24 | 78.69±1.93 | 69.57±8.35 |
| text-davinci-003 | 69.82±5.31 | 98.26±1.88 | 62.02±12.80 | 69.62±4.56 | 69.50±5.41 |
| gpt-3.5-turbo | 68.25±1.66 | 99.00±0.50 | 73.04±4.82 | 81.74±3.27 | 69.75±2.74 |
| *3-shot* | | | | | |
| code-davinci-002 | **73.00±1.00** | **100.00±0.00** | 80.67±4.51 | **91.00±5.57** | **84.48±0.18** |
| text-davinci-001 | 56.05±2.91 | 98.42±1.12 | 50.00±9.74 | 75.11±5.74 | 53.80±4.41 |
| text-davinci-002 | 73.14±2.60 | 96.10±6.53 | 66.45±5.80 | 85.86±9.69 | 72.70±3.57 |
| text-davinci-003 | 67.99±7.34 | 97.51±1.63 | 60.55±17.15 | 68.78±7.42 | 68.50±10.26 |
| gpt-3.5-turbo | 69.49±1.62 | 97.92±1.50 | 75.37±4.87 | 78.35±4.59 | 70.34±1.23 |

Table 20: Performance and robustness test results (accuracy) of different models in zero-shot and few-shot scenarios on **QQP** dataset.

| Model | NumWord<br># 2670 samples | | SwapAnt<br># 883 samples | | all<br># 40430 samples |
|---|---|---|---|---|---|
| | ori | trans | ori | trans | ori |
| *0-shot* | | | | | |
| code-davinci-002 | 7.95±13.78 | 0.00±0.00 | 32.14±55.67 | 0.68±1.18 | 37.67±9.61 |
| text-davinci-001 | 35.37±2.99 | 6.07±9.48 | 75.65±23.45 | 14.01±19.48 | 36.40±5.82 |
| text-davinci-002 | 68.07±5.70 | 25.00±19.56 | **85.32±13.47** | 28.43±20.73 | 63.00±5.69 |
| text-davinci-003 | 79.85±1.37 | **73.22±19.84** | 60.44±8.78 | **65.31±5.15** | **81.03±0.67** |
| gpt-3.5-turbo | **80.09±3.15** | 65.67±27.30 | 78.55±8.79 | 64.25±17.48 | 79.23±2.43 |
| *1-shot* | | | | | |
| code-davinci-002 | 71.63±4.88 | 58.64±33.60 | 34.22±32.00 | 53.89±35.10 | 68.33±4.04 |
| text-davinci-001 | 66.17±3.43 | 86.47±11.10 | 26.61±23.80 | **90.79±11.85** | 66.50±3.35 |
| text-davinci-002 | 79.50±7.11 | **90.17±24.47** | 47.41±12.95 | 67.87±6.12 | 77.70±3.05 |
| text-davinci-003 | 79.81±1.69 | 73.98±20.42 | 63.42±9.92 | 51.34±7.89 | 80.93±1.91 |
| gpt-3.5-turbo | 81.78±1.47 | 67.67±18.56 | **72.27±14.08** | 76.62±12.30 | 79.21±1.79 |
| *3-shot* | | | | | |
| code-davinci-002 | 78.12±8.61 | 71.78±11.74 | 42.89±15.38 | 75.95±11.33 | 79.00±2.65 |
| text-davinci-001 | 72.60±6.48 | 72.90±12.68 | 67.01±6.18 | 76.48±9.44 | 65.17±5.29 |
| text-davinci-002 | 82.60±6.87 | **87.13±19.29** | 57.53±14.28 | 75.27±6.91 | 80.30±1.15 |
| text-davinci-003 | 83.35±0.14 | 71.06±15.50 | **71.46±9.13** | 56.21±8.66 | **82.97±1.72** |
| gpt-3.5-turbo | **83.47±1.46** | 70.67±19.86 | 68.88±9.87 | 73.35±17.41 | 80.69±0.78 |

We evaluate the SM ability of GPT series models using MRPC and QQP datasets in both zero-shot and few-shot scenarios.

Our findings indicate that **in the zero-shot setting, text-davinci-003 and gpt-3.5-turbo have better performance than others, while code-davinci-002 and text-davinci-001 perform pooly, as shown in Table 19 and Table 20.** We also observe that 1)*NumWord* induces a significant drop in average performance, as it requires the model to perform numerical reasoning for correct semantic inference. 2)

*SwapAnt* results in up to a 61.64% drop in average performance, indicating that the models struggle with the semantic contradiction expressed by antonyms between premise-hypothesis pairs.

**In few-shot scenarios, we see significant improvement in both performance and robustness of the GPT series models.** pecifically, code-davinci-002 exhibits a significant ability in 3-shot settings in MRPC datasets and is more sensitive to numerical inputs. In QQP datasets, as the number of samples in the prompt increases, the performance difference between models decreases.

### 4.2.9 The Winograd Schema Challenge

Table 21: Performance and robustness test results (accuracy) of GPT series models in zero-shot and few-shot scenarios on **WSC** dataset.

| Model | all<br># 570 samples | AddSentences<br># 570 samples | InsertRelativeClause<br># 566 samples | SwapNames<br># 566 samples | SwitchVoice<br># 440 samples | SwapGender<br># 310 samples |
|---|---|---|---|---|---|---|
| | | | *0-shot* | | | |
| code-davinci-002 | 50.67±0.58 | 49.67±0.58 | 50.67±1.15 | 51.00±1.00 | 50.67±0.58 | 50.84±1.46 |
| text-davinci-001 | 52.05±1.14 | 53.22±1.60 | 50.94±1.37 | 51.12±1.41 | 51.14±0.82 | 52.04±0.38 |
| text-davinci-002 | 61.46±1.57 | 64.09±1.00 | 56.89±1.23 | 59.84±1.77 | 59.47±1.77 | 60.97±2.01 |
| text-davinci-003 | 62.05±0.57 | 65.32±1.77 | **59.83±0.51** | 60.48±0.74 | 59.39±0.92 | 63.01±1.30 |
| gpt-3.5-turbo | **66.05±2.31** | 70.56±10.15 | 59.20±3.85 | **64.83±2.31** | **62.55±2.21** | **63.44±2.27** |
| | | | *1-shot* | | | |
| code-davinci-002 | 58.00±3.61 | 58.00±2.65 | 56.33±0.58 | 56.33±4.93 | 54.67±3.51 | 58.33±4.16 |
| text-davinci-001 | 50.41±0.56 | 53.39±2.78 | 50.18±0.36 | 49.76±0.10 | 50.23±0.39 | 50.65±0.65 |
| text-davinci-002 | 60.94±2.03 | 65.20±3.62 | 58.54±2.58 | 59.54±2.46 | 57.80±1.26 | 61.61±1.48 |
| text-davinci-003 | 61.40±1.37 | 62.98±0.63 | 58.18±0.57 | 58.42±0.44 | 57.35±0.13 | 61.72±0.49 |
| gpt-3.5-turbo | 59.77±1.06 | 60.76±1.06 | 59.01±1.91 | 59.13±1.54 | 56.67±0.95 | 59.89±2.27 |
| | | | *3-shot* | | | |
| code-davinci-002 | 57.00±1.00 | 63.33±3.79 | 53.00±1.00 | 58.00±1.73 | 58.33±4.04 | 58.33±2.08 |
| text-davinci-001 | 51.34±0.97 | 54.79±2.11 | 51.47±0.84 | 51.65±0.80 | 50.83±1.84 | 50.43±1.83 |
| text-davinci-002 | 62.22±1.47 | 64.91±0.88 | **59.07±2.12** | **62.07±0.89** | 58.18±1.42 | 61.07±2.44 |
| text-davinci-003 | **62.75±1.13** | **65.38±0.53** | 57.95±0.47 | 61.37±0.80 | **59.62±0.57** | **63.23±0.65** |
| gpt-3.5-turbo | 58.48±3.01 | 59.82±2.13 | 56.18±2.32 | 60.01±3.18 | 58.41±2.31 | 58.39±4.77 |

We conduct experiments on the WSC273 dataset for the WSC task and report the accuracy in Table 21. **In the zero-shot scenario, gpt-3.5-turbo consistently achieves the best performance, as shown in the table.** The performance of text-davinci-003 and the text-davinci-002 is close to it, while text-davinci-001 and code-davinci-002 lag behind. **In the few-shot scene, we observe that various deformations achieve the best performance in text-davinci-003 or text-davinci-002 set in three-shot, while text-davinci-001 shows the worst performance.** It is noteworthy that in the WSC dataset, the model's performance does not always increase with the number of examples in the prompt. In fact, the performance of the model declines in the transition from zero-shot to one-shot, and there is no obvious trend in robustness.

## 5 Conclusion

In this paper, we comprehensively analyze the capabilities of six GPT series models, including GPT-3 and GPT-3.5, by evaluating their performance and robustness on 21 datasets across nine NLU tasks. Our findings reveal that the evolution of GPT series models does not necessarily lead to universal improvements across all NLU tasks, which is influenced by the training strategy employed and the specific characteristics of each task. Moreover, we observe that despite the improved performance of the models, their robustness does not show significant enhancements, which warrants further investigation. We hope that our study will offer new insights to future work on how to balance the model's task-solving ability with its user-friendly response capabilities, as well as on how to improve its robustness while enhancing its performance.

## 6 Limitations

In this paper, we systematically analyze the GPT-3 and GPT-3.5 series and summarize some findings and conclusions. However, we acknowledge that there are some limitations. Firstly, we do not use the full dataset for testing some models due to the OpenAI API limiting the rate of accesses, but this does not affect the overall trend analysis. Secondly, OpenAI releases GPT-4 during our study and notes that it has more powerful capabilities. Unfortunately, the GPT-4 API has not been made available yet, which has made it difficult for us to test whether GPT-4 addresses some of the issues with the previous model. Investigating this will be a critical area for future research.

## References

Tom B. Brown, Benjamin Mann, Nick Ryder, Melanie Subbiah, Jared Kaplan, Prafulla Dhariwal, Arvind Neelakantan, Pranav Shyam, Girish Sastry, Amanda Askell, Sandhini Agarwal, Ariel Herbert-Voss, Gretchen Krueger, Tom Henighan, Rewon Child, Aditya Ramesh, Daniel M. Ziegler, Jeffrey Wu, Clemens Winter, Christopher Hesse, Mark Chen, Eric Sigler, Mateusz Litwin, Scott Gray, Benjamin Chess, Jack Clark, Christopher Berner, Sam McCandlish, Alec Radford, Ilya Sutskever, and Dario Amodei. 2020. Language models are few-shot learners. In *Advances in Neural Information Processing Systems 33: Annual Conference on Neural Information Processing Systems 2020, NeurIPS 2020, December 6-12, 2020, virtual*.

Mark Chen, Jerry Tworek, Heewoo Jun, Qiming Yuan, Henrique Pondé de Oliveira Pinto, Jared Kaplan, Harrison Edwards, Yuri Burda, Nicholas Joseph, Greg Brockman, Alex Ray, Raul Puri, Gretchen Krueger, Michael Petrov, Heidy Khlaaf, Girish Sastry, Pamela Mishkin, Brooke Chan, Scott Gray, Nick Ryder, Mikhail Pavlov, Alethea Power, Lukasz Kaiser, Mohammad Bavarian, Clemens Winter, Philippe Tillet, Felipe Petroski Such, Dave Cummings, Matthias Plappert, Fotios Chantzis, Elizabeth Barnes, Ariel Herbert-Voss, William Hebgen Guss, Alex Nichol, Alex Paino, Nikolas Tezak, Jie Tang, Igor Babuschkin, Suchir Balaji, Shantanu Jain, William Saunders, Christopher Hesse, Andrew N. Carr, Jan Leike, Joshua Achiam, Vedant Misra, Evan Morikawa, Alec Radford, Matthew Knight, Miles Brundage, Mira Murati, Katie Mayer, Peter Welinder, Bob McGrew, Dario Amodei, Sam McCandlish, Ilya Sutskever, and Wojciech Zaremba. 2021. Evaluating large language models trained on code. *CoRR*, abs/2107.03374.

Xuanting Chen, Junjie Ye, Can Zu, Nuo Xu, Rui Zheng, Minlong Peng, Jie Zhou, Tao Gui, Qi Zhang, and Xuanjing Huang. 2023. How robust is gpt-3.5 to predecessors? a comprehensive study on language understanding tasks. *ArXiv*, abs/2303.00293.

Aakanksha Chowdhery, Sharan Narang, Jacob Devlin, Maarten Bosma, Gaurav Mishra, Adam Roberts, Paul Barham, Hyung Won Chung, Charles Sutton, Sebastian Gehrmann, Parker Schuh, Kensen Shi, Sasha Tsvyashchenko, Joshua Maynez, Abhishek Rao, Parker Barnes, Yi Tay, Noam Shazeer, Vinodkumar Prabhakaran, Emily Reif, Nan Du, Ben Hutchinson, Reiner Pope, James Bradbury, Jacob Austin, Michael Isard, Guy Gur-Ari, Pengcheng Yin, Toju Duke, Anselm Levskaya, Sanjay Ghemawat, Sunipa Dev, Henryk Michalewski, Xavier Garcia, Vedant Misra, Kevin Robinson, Liam Fedus, Denny Zhou, Daphne Ippolito, David Luan, Hyeontaek Lim, Barret Zoph, Alexander Spiridonov, Ryan Sepassi, David Dohan, Shivani Agrawal, Mark Omernick, Andrew M. Dai, Thanumalayan Sankaranarayana Pillai, Marie Pellat, Aitor Lewkowycz, Erica Moreira, Rewon Child, Oleksandr Polozov, Katherine Lee, Zongwei Zhou, Xuezhi Wang, Brennan Saeta, Mark Diaz, Orhan Firat, Michele Catasta, Jason Wei, Kathy Meier-Hellstern, Douglas Eck, Jeff Dean, Slav Petrov, and Noah Fiedel. 2022. Palm: Scaling language modeling with pathways. *CoRR*, abs/2204.02311.

Paul F. Christiano, Jan Leike, Tom B. Brown, Miljan Martic, Shane Legg, and Dario Amodei. 2017. Deep reinforcement learning from human preferences. In *Advances in Neural Information Processing Systems 30: Annual Conference on Neural Information Processing Systems 2017, December 4-9, 2017, Long Beach, CA, USA*, pages 4299–4307.

Bill Dolan and Chris Brockett. 2005. Automatically constructing a corpus of sentential paraphrases. In *Third International Workshop on Paraphrasing (IWP2005)*.

Kevin Gimpel, Nathan Schneider, Brendan O'Connor, Dipanjan Das, Daniel Mills, Jacob Eisenstein, Michael Heilman, Dani Yogatama, Jeffrey Flanigan, and Noah A Smith. 2010. Part-of-speech tagging for twitter: Annotation, features, and experiments. Technical report, Carnegie-Mellon Univ Pittsburgh Pa School of Computer Science.

Tao Gui, Xiao Wang, Qi Zhang, Qin Liu, Yicheng Zou, Xin Zhou, Rui Zheng, Chong Zhang, Qinzhuo Wu, Jiacheng Ye, et al. 2021. Textflint: Unified multilingual robustness evaluation toolkit for natural language processing. *arXiv preprint arXiv:2103.11441*.

Amr Hendy, Mohamed Gomaa Abdelrehim, Amr Sharaf, Vikas Raunak, Mohamed Gabr, Hitokazu Matsushita, Young Jin Kim, Mohamed Afify, and Hany Hassan Awadalla. 2023. How good are gpt models at machine translation? a comprehensive evaluation. *ArXiv*, abs/2302.09210.

Jan Koco'n, Igor Cichecki, Oliwier Kaszyca, Mateusz Kochanek, Dominika Szydlo, Joanna Baran, Julita Bielaniewicz, Marcin Gruza, Arkadiusz Janz, Kamil Kanclerz, Anna Koco'n, Bartlomiej Koptyra, Wiktoria Mieleszczenko-Kowszewicz, P. Milkowski, Marcin Oleksy, Maciej Piasecki, Lukasz Radli'nski, Konrad Wojtasik, Stanislaw Wo'zniak, and Przemyslaw Kazienko. 2023. Chatgpt: Jack of all trades, master of none. *ArXiv*, abs/2302.10724.

Hector Levesque, Ernest Davis, and Leora Morgenstern. 2012. The winograd schema challenge. In *Thirteenth International Conference on the Principles of Knowledge Representation and Reasoning*. Citeseer.

Gina-Anne Levow. 2006. The third international chinese language processing bakeoff: Word segmentation and named entity recognition. In *Proceedings of the Fifth SIGHAN workshop on Chinese language processing*, pages 108–117.

Andrew Maas, Raymond E Daly, Peter T Pham, Dan Huang, Andrew Y Ng, and Christopher Potts. 2011. Learning word vectors for sentiment analysis. In *Proceedings of the 49th annual meeting of the association for computational linguistics: Human language technologies*, pages 142–150.

Mitchell P. Marcus, Beatrice Santorini, and Mary Ann Marcinkiewicz. 1993. Building a large annotated corpus of English: The Penn Treebank. *Computational Linguistics*, 19(2):313–330.

Long Ouyang, Jeff Wu, Xu Jiang, Diogo Almeida, Carroll L Wainwright, Pamela Mishkin, Chong Zhang, Sandhini Agarwal, Katarina Slama, Alex Ray, et al. 2022. Training language models to follow instructions with human feedback. *arXiv preprint arXiv:2203.02155*.

Maria Pontiki, Dimitris Galanis, John Pavlopoulos, Harris Papageorgiou, Ion Androutsopoulos, and Suresh Manandhar. 2014. SemEval-2014 task 4: Aspect based sentiment analysis. In *Proceedings of the 8th International Workshop on Semantic Evaluation (SemEval 2014)*, pages 27–35, Dublin, Ireland. Association for Computational Linguistics.

Chengwei Qin, Aston Zhang, Zhuosheng Zhang, Jiaao Chen, Michihiro Yasunaga, and Diyi Yang. 2023. Is chatgpt a general-purpose natural language processing task solver? *arXiv preprint arXiv:2302.06476*.

Pranav Rajpurkar, Robin Jia, and Percy Liang. 2018. Know what you don't know: Unanswerable questions for squad. *CoRR*, abs/1806.03822.

Pranav Rajpurkar, Jian Zhang, Konstantin Lopyrev, and Percy Liang. 2016. Squad: 100, 000+ questions for machine comprehension of text. *CoRR*, abs/1606.05250.

Erik Tjong Kim Sang and Fien De Meulder. 2003. Introduction to the conll-2003 shared task: language-independent named entity recognition. *North American Chapter of the Association for Computational Linguistics*.

Zhiguo Wang, Wael Hamza, and Radu Florian. 2017. Bilateral multi-perspective matching for natural language sentences. *CoRR*, abs/1702.03814.

Jason Wei, Maarten Bosma, Vincent Y. Zhao, Kelvin Guu, Adams Wei Yu, Brian Lester, Nan Du, Andrew M. Dai, and Quoc V. Le. 2022. Finetuned language models are zero-shot learners. In *The *Tenth International Conference on Learning Representations, ICLR 2022, Virtual Event, April 25-29, 2022*. OpenReview.net.

Ralph Weischedel, Martha Palmer, Mitchell Marcus, Eduard Hovy, Sameer Pradhan, Lance Ramshaw, Nianwen Xue, Ann Taylor, Jeff Kaufman, Michelle Franchini, et al. 2013. Ontonotes release 5.0 ldc2013t19. *Linguistic Data Consortium, Philadelphia, PA*, 23.

Adina Williams, Nikita Nangia, and Samuel R. Bowman. 2017. A broad-coverage challenge corpus for sentence understanding through inference. *CoRR*, abs/1704.05426.

Xianjun Yang, Yan Li, Xinlu Zhang, Haifeng Chen, and Wei Cheng. 2023. Exploring the limits of chatgpt for query or aspect-based text summarization. *ArXiv*, abs/2302.08081.

Lining Zhang, M. Wang, Liben Chen, and Wenxin Zhang. 2022a. Probing gpt-3's linguistic knowledge on semantic tasks. In *BlackboxNLP Workshop on Analyzing and Interpreting Neural Networks for NLP*.

Susan Zhang, Stephen Roller, Naman Goyal, Mikel Artetxe, Moya Chen, Shuohui Chen, Christopher Dewan, Mona T. Diab, Xian Li, Xi Victoria Lin, Todor Mihaylov, Myle Ott, Sam Shleifer, Kurt Shuster, Daniel Simig, Punit Singh Koura, Anjali Sridhar, Tianlu Wang, and Luke Zettlemoyer. 2022b. OPT: open pre-trained transformer language models. *CoRR*, abs/2205.01068.

Yuhao Zhang, Victor Zhong, Danqi Chen, Gabor Angeli, and Christopher D. Manning. 2017. Position-aware attention and supervised data improve slot filling. In *Proceedings of the 2017 Conference on Empirical Methods in Natural Language Processing, EMNLP 2017, Copenhagen, Denmark, September 9-11, 2017*, pages 35–45. Association for Computational Linguistics.

## A Performance Details

### A.1 Performance of Different Models in the Zero-shot Scenario

Table 22: Performance of different models in the zero-shot scenario. With the exception of davinci, "-" indicates that the non-analyzable rate exceeds the threshold and counts as not completing the specified task.

| Task | Dataset | davinci | code-davinci-002 | text-davinci-001 | text-davinci-002 | text-davinci-003 | gpt-3.5-turbo |
|---|---|---|---|---|---|---|---|
| Aspect-based Sentiment Analysis | laptop | 79.00 | 90.72 | 83.91 | 86.48 | 84.12 | 85.11 |
|  | restaurant | 10.00 | 93.00 | 89.73 | 91.26 | 88.78 | 91.02 |
| Machine Reading Comprehension | SQuAD1.1 | 69.13 | 83.44 | 86.17 | 89.01 | 78.22 | 65.71 |
|  | SQuAD2.0 | 51.58 | 81.76 | 76.47 | 79.66 | 76.46 | 65.53 |
| Named Entity Recognition | ACE 2005 | 1.44 | 23.32 | 19.39 | 33.64 | 34.88 | 34.84 |
|  | CoNLL 2003 | 10.89 | 65.95 | 20.20 | 55.46 | 51.83 | 44.81 |
|  | OntoNotes v5 | 0.00 | 0.00 | 0.29 | 2.21 | 6.56 | 14.47 |
|  | HONOR | 9.79 | 44.61 | - | 51.57 | 49.69 | 51.50 |
|  | MSRANER | 17.50 | 10.09 | - | 14.96 | 19.48 | 16.43 |
|  | OntoNote4NER | 0.01 | 15.94 | - | - | 31.92 | 11.44 |
| Natural Language Inference | MNLI-m | 30.00 | 48.00 | 43.48 | 45.52 | 63.66 | 67.87 |
|  | MNLI-mm | 31.00 | 50.00 | 45.74 | 42.65 | 64.38 | 68.13 |
|  | SNLI | 31.00 | 57.00 | 46.53 | 38.10 | 68.56 | 64.66 |
| Part-of-speech Tagging | WSJ | 0.00 | 22.87 | - | 70.31 | 88.45 | - |
|  | Daily547 | 0.20 | 39.89 | - | 48.42 | 64.69 | - |
|  | PKU-SEGPOS | 0.00 | 50.80 | - | 33.46 | 66.05 | 57.36 |
| Relation Extraction | Tacred | 2.00 | 15.00 | 10.87 | 12.17 | 21.90 | 15.70 |
| Sentiment Classification | IMDB | 78.00 | 94.00 | 93.18 | 90.60 | 91.76 | 91.13 |
| Semantic Matching | MRPC | 32.00 | 73.51 | 43.80 | 75.20 | 74.50 | 73.74 |
|  | QQP | 32.00 | 36.00 | 43.10 | 69.50 | 81.20 | 77.06 |
| The Winograd Schema Challenge | WSC273 | 50.00 | 50.00 | 52.11 | 59.65 | 61.40 | 66.60 |

## A.2 Analyzability Rate and Performance of davinci in All Datasets

Table 23: Analyzability rate and performance of davinci in zero-shot and few-shot scenarios. We manually calculated the evaluation results and considered the non-analyzable results as wrong answers.

| Task | Dataset | 0-shot | | 3-shot | |
| --- | --- | --- | --- | --- | --- |
| | | Analyzable rate | Evaluation results | Analyzable rate | Evaluation results |
| Aspect-based Sentiment Analysis | SemEval2014-Laptop | 86.00 | 79.00 | 100.00 | 96.00 |
| | SemEval2014-Restaurant | 13.00 | 10.00 | 100.00 | 99.00 |
| Machine Reading Comprehension | SQuAD1.1 | 88.00 | 69.13 (F1) | 100.00 | 87.07 (F1) |
| | SQuAD2.0 | 89 | 51.58 (F1) | 100.00 | 78.57 (F1) |
| Named Entity Recognition | ACE 2005 | 2.00 | 1.44 | 100.00 | 33.03 |
| | CoNLL 2003 | 33.00 | 10.89 | 72.00 | 46.61 |
| | OntoNotes v5 | 5.00 | 0.00 | 100.00 | 0.00 |
| | HONOR | 20.00 | 9.79 | 88.00 | 39.50 |
| | MSRANER | 72.00 | 17.50 | 100.00 | 43.31 |
| | OntoNote4NER | 2.00 | 0.01 | 100.00 | 30.68 |
| Natural Language Inference | MNLI-m | 100.00 | 30.00 | 100.00 | 34.00 |
| | MNLI-mm | 100.00 | 31.00 | 100.00 | 35.00 |
| | SNLI | 86.00 | 31.00 | 100.00 | 35.00 |
| Part-of-speech Tagging | Daily547 | 3.00 | 0.20 | 95.00 | 48.63 |
| | WSJ | 0.00 | 0.00 | 90.00 | 47.12 |
| | PKU-SEGPOS | 0.00 | 0.00 | 100.00 | 34.71 |
| Relation Extraction | Tacred | 15.00 | 2.00 | 100.00 | 8.00 |
| Sentiment Classification | IMDB | 94.00 | 78.00 | 87.00 | 85.00 |
| Semantic Matching | MRPC | 100.00 | 32.00 | 100.00 | 68.00 |
| | QQP | 100.00 | 32.00 | 100.00 | 70.00 |
| The Winograd Schema Challenge | WSC273 | 100.00 | 50.00 | 100.00 | 50.00 |

## A.3 Analyzability Comparison of davinci

Table 24: Analyzability comparison of davinci in zero-shot scenario. The "w/o 'Answer'" means there is no "Answer" added at the end of prompt in zero-shot setting, which decreases the analyzability of davinci model.

| Task | Dataset | w/o "Answer" | w/ "Answer" |
| --- | --- | --- | --- |
| Aspect-based Sentiment Analysis | SemEval2014-Laptop | 13.00 | 87.00 |
| | SemEval2014-Restaurant | 14.00 | 13.00 |
| Natural Language Inference | MNLI-m | 71.00 | 100.00 |
| | MNLI-mm | 59.00 | 100.00 |
| | SNLI | 0.00 | 86.00 |
| Sentiment Classification | IMDB | 95.00 | 94.00 |
| Semantic Matching | MRPC | 0.00 | 100.00 |
| | QQP | 33.00 | 100.00 |

## B Prompts

For each dataset, we designed three prompts in the 0/1/3-shot scenario, respectively. Since 3-shot just adds more examples in the prompt compared to 1-shot, we list the prompts we use for each dataset in Table 25 to Table 45 for zero-shot and 1-shot.

Table 25: 0/1-shot prompts for SemEval2014-Laptop dataset. The "{aspect}" should be replaced by the aspect to be analyzed, and the "{sentence}" should be replaced by a sentence.

| # Shot | Prompts |
| --- | --- |
| zero-shot | Analyze the sentiment towards the '{aspect}' of '{sentence}' and determine if it is positive, negative, or neutral. // Answer:<br><br>What is the sentiment towards '{sentence}' in terms of '{aspect}'? Are they viewed positively, negatively, or neutrally? // Answer:<br><br>'{sentence}' Express your sentiment towards the aspect of '{aspect}' using positive, negative, or neutral. // Answer: |
| 1-shot | Analyze the sentiment towards the 'BIOS' of 'But sadly the replacement froze up while updating the BIOS again and shut down and would not turn back on.' and determine if it is positive, negative, or neutral. // Answer: negative // Analyze the sentiment towards the '{aspect}' of '{sentence}' and determine if it is positive, negative, or neutral. // Answer:<br><br>What is the sentiment towards 'But sadly the replacement froze-up while updating the BIOS again and shut down and would not turn back on.' in terms of 'BIOS'? Are they viewed positively, negatively, or neutrally? // Answer: negative // What is the sentiment towards '{sentence}' in terms of '{aspect}'? Are they viewed positively, negatively, or neutrally? // Answer:<br><br>'But sadly the replacement froze-up while updating the BIOS again and shut down and would not turn back on.' Express your sentiment towards the aspect of 'BIOS' using positive, negative, or neutral. // Answer: negative // '{sentence}' Express your sentiment towards the aspect of '{aspect}' using positive, negative, or neutral. // Answer: |

Table 26: 0/1-shot prompts for SemEval2014-Restaurant dataset. The "{aspect}" should be replaced by the aspect to be analyzed, and the "{sentence}" should be replaced by the sentence.

| # Shot | Prompts |
| --- | --- |
| zero-shot | Analyze the sentiment towards the '{aspect}' of '{sentence}' and determine if it is positive, negative, or neutral. // Answer:<br><br>What is the sentiment towards '{sentence}' in terms of '{aspect}'? Are they viewed positively, negatively, or neutrally? // Answer:<br><br>'{sentence}' Express your sentiment towards the aspect of '{aspect}' using positive, negative, or neutral. // Answer: |
| 1-shot | Analyze the sentiment towards the 'dishes' of 'The food is good, especially their more basic dishes, and the drinks are delicious.' and determine if it is positive, negative, or neutral. // Answer: positive // Analyze the sentiment towards the '{aspect}' of '{sentence}' and determine if it is positive, negative, or neutral. // Answer:<br><br>What is the sentiment towards 'The food is good, especially their more basic dishes, and the drinks are delicious.' in terms of 'dishes'? Are they viewed positively, negatively, or neutrally? // Answer: positive // What is the sentiment towards '{sentence}' in terms of '{aspect}'? Are they viewed positively, negatively, or neutrally? // Answer:<br><br>'The food is good, especially their more basic dishes, and the drinks are delicious.' Express your sentiment towards the aspect of 'dishes' using positive, negative, or neutral. // Answer: positive // '{sentence}' Express your sentiment towards the aspect of '{aspect}' using positive, negative, or neutral. // Answer: |

Table 27: 0/1-shot prompts for SQuAD1.0 dataset. The "{context}" should be replaced by passage, the "{question}" should be replaced by question.

| # Shot | Prompts |
| --- | --- |
| zero-shot | Passage:{context} // Question: {question} // Referring to the passage above, the correct answer to the given question is // Answer: |
| | Refer to the passage below and answer the following question: // Passage: {context} // Question: {question} // Answer: |
| | Passage: {context} // Question: {question} // Answer: |
| 1-shot | Passage: 'Architecturally, the school has a Catholic character. Atop the Main Building's gold dome is a golden statue of the Virgin Mary. Immediately in front of the Main Building and facing it, is a copper statue of Christ with arms upraised with the legend "Venite Ad Me Omnes". Next to the Main Building is the Basilica of the Sacred Heart. Immediately behind the basilica is the Grotto, a Marian place of prayer and reflection. It is a replica of the grotto at Lourdes, France where the Virgin Mary reputedly appeared to Saint Bernadette Soubirous in 1858. At the end of the main drive (and in a direct line that connects through 3 statues and the Gold Dome), is a simple, modern stone statue of Mary. ' // Question: 'To whom did the Virgin Mary allegedly appear in 1858 in Lourdes France?' // Referring to the passage above, the correct answer to the given question is // Answer: Saint Bernadette Soubirous // Passage:'{context}' // Question: '{question}' // Referring to the passage above, the correct answer to the given question is |
| | Refer to the passage below and answer the following question: // Passage: 'Architecturally, the school has a Catholic character. Atop the Main Building's gold dome is a golden statue of the Virgin Mary. Immediately in front of the Main Building and facing it, is a copper statue of Christ with arms upraised with the legend "Venite Ad Me Omnes". Next to the Main Building is the Basilica of the Sacred Heart. Immediately behind the basilica is the Grotto, a Marian place of prayer and reflection. It is a replica of the grotto at Lourdes, France where the Virgin Mary reputedly appeared to Saint Bernadette Soubirous in 1858. At the end of the main drive (and in a direct line that connects through 3 statues and the Gold Dome), is a simple, modern stone statue of Mary. ' // Question: 'To whom did the Virgin Mary allegedly appear in 1858 in Lourdes France?' // Answer: Saint Bernadette Soubirous // Refer to the passage below and answer the following question: // Passage: '{context}' // Question: '{question}' // Answer: |
| | Passage: 'Architecturally, the school has a Catholic character. Atop the Main Building's gold dome is a golden statue of the Virgin Mary. Immediately in front of the Main Building and facing it, is a copper statue of Christ with arms upraised with the legend "Venite Ad Me Omnes". Next to the Main Building is the Basilica of the Sacred Heart. Immediately behind the basilica is the Grotto, a Marian place of prayer and reflection. It is a replica of the grotto at Lourdes, France where the Virgin Mary reputedly appeared to Saint Bernadette Soubirous in 1858. At the end of the main drive (and in a direct line that connects through 3 statues and the Gold Dome), is a simple, modern stone statue of Mary. ' // Question: 'To whom did the Virgin Mary allegedly appear in 1858 in Lourdes France?' // Answer: Saint Bernadette Soubirous // Passage: '{context}' // Question: '{question}' // Answer: |

Table 28: 0/1-shot prompts for SQuAD2.0 dataset. The "{context}" should be repaced by passage, and the "{question}" should be replaced by question.

| # Shot | Prompts |
| --- | --- |
| zero-shot | Passage:{context} // Question: {question} // Referring to the passage above, the correct answer to the given question is // Answer: |
| | Refer to the passage below and answer the following question: // Passage: {context} // Question: {question} // Answer: |
| | Passage: {context} // Question: {question} // Answer: |
| 1-shot | Passage: 'Architecturally, the school has a Catholic character. Atop the Main Building's gold dome is a golden statue of the Virgin Mary. Immediately in front of the Main Building and facing it, is a copper statue of Christ with arms upraised with the legend "Venite Ad Me Omnes". Next to the Main Building is the Basilica of the Sacred Heart. Immediately behind the basilica is the Grotto, a Marian place of prayer and reflection. It is a replica of the grotto at Lourdes, France where the Virgin Mary reputedly appeared to Saint Bernadette Soubirous in 1858. At the end of the main drive (and in a direct line that connects through 3 statues and the Gold Dome), is a simple, modern stone statue of Mary. ' // Question: 'To whom did the Virgin Mary allegedly appear in 1858 in Lourdes France?' // Referring to the passage above, the correct answer to the given question is // Answer: Saint Bernadette Soubirous // Passage:'{context}' // Question: '{question}' // Referring to the passage above, the correct answer to the given question is |
| | Refer to the passage below and answer the following question: // Passage: 'Architecturally, the school has a Catholic character. Atop the Main Building's gold dome is a golden statue of the Virgin Mary. Immediately in front of the Main Building and facing it, is a copper statue of Christ with arms upraised with the legend "Venite Ad Me Omnes". Next to the Main Building is the Basilica of the Sacred Heart. Immediately behind the basilica is the Grotto, a Marian place of prayer and reflection. It is a replica of the grotto at Lourdes, France where the Virgin Mary reputedly appeared to Saint Bernadette Soubirous in 1858. At the end of the main drive (and in a direct line that connects through 3 statues and the Gold Dome), is a simple, modern stone statue of Mary. ' // Question: 'To whom did the Virgin Mary allegedly appear in 1858 in Lourdes France?' // Answer: Saint Bernadette Soubirous // Refer to the passage below and answer the following question: // Passage: '{context}' // Question: '{question}' // Answer: |
| | Passage: 'Architecturally, the school has a Catholic character. Atop the Main Building's gold dome is a golden statue of the Virgin Mary. Immediately in front of the Main Building and facing it, is a copper statue of Christ with arms upraised with the legend "Venite Ad Me Omnes". Next to the Main Building is the Basilica of the Sacred Heart. Immediately behind the basilica is the Grotto, a Marian place of prayer and reflection. It is a replica of the grotto at Lourdes, France where the Virgin Mary reputedly appeared to Saint Bernadette Soubirous in 1858. At the end of the main drive (and in a direct line that connects through 3 statues and the Gold Dome), is a simple, modern stone statue of Mary. ' // Question: 'To whom did the Virgin Mary allegedly appear in 1858 in Lourdes France?' // Answer: Saint Bernadette Soubirous // Passage: '{context}' // Question: '{question}' // Answer: |

Table 29: 0/1-shot prompts for ACE 2005 dataset. The "{text}" should be replaced by sentence.

| # Shot | Prompts |
| --- | --- |
| zero-shot | Please identify Organization, Person, Geo-political Entity, Facility, Location, Vehicle and Weapon Entity from the given text, output using the format as "Entity: Organization: None\|Person: Word1\|Geo-political Entity: None\|Facility: Word2\|Location: Word3, Word4\|Vehicle: None\|Weapon: None" // Text: {text} // Entity:<br><br>Please list all Organization, Person, Geo-political Entity, Facility, Location, Vehicle and Weapon Entity in the given text, output using the format as "Entity: Organization: None\|Person: Word1\|Geo-political Entity: None\|Facility: Word2\|Location: Word3, Word4\|Vehicle: None\|Weapon: None" // Text: {text} // Entity:<br><br>Extract all Organization, Person, Geo-political Entity, Facility, Location, Vehicle and Weapon Entity from the given text, output using the format as "Entity: Organization: None\|Person: Word1\|Geo-political Entity: None\|Facility: Word2\|Location: Word3, Word4\|Vehicle: None\|Weapon: None" // Text: {text} // Entity: |
| 1-shot | Please identify Organization, Person, Geo-political Entity, Facility, Location, Vehicle and Weapon Entity from the given text // Text: thank you , paula . // Entity: Organization: None\|Person: you, paula\|Geo-political Entity: None\|Facility: None\|Location: None\|Vehicle: None\|Weapon: None // Text: {text} // Entity:<br><br>Please list all Organization, Person, Geo-political Entity, Facility, Location, Vehicle and Weapon Entity in the given text // Text: thank you , paula . // Entity: Organization: None\|Person: you, paula\|Geo-political Entity: None\|Facility: None\|Location: None\|Vehicle: None\|Weapon: None // Text: {text} // Entity:<br><br>Extract all Organization, Person, Geo-political Entity, Facility, Location, Vehicle and Weapon Entity from the given text // Text: thank you , paula . // Entity: Organization: None\|Person: you, paula\|Geo-political Entity: None\|Facility: None\|Location: None\|Vehicle: None\|Weapon: None // Text: {text} // Entity: |

Table 30: 0/1-shot prompts for CoNLL 2003 dataset. The "{text}" should be replaced by text.

| # Shot | Prompts |
| --- | --- |
| zero-shot | Please identify Organization, Person, Location and Miscellaneous Entity from the given text, output using the format as "Entity: Organization: None\|Person: None\|Location: Word1, Word2\|Miscellaneous: Word3" // Text: {text} // Entity:<br><br>Please list all Organization, Person, Location and Miscellaneous Entity in the given text, output using the format as "Entity: Organization: None\|Person: None\|Location: Word1, Word2\|Miscellaneous: Word3" // Text: {text} // Entity:<br><br>Extract all Organization, Person, Location and Miscellaneous Entity from the given text, output using the format as "Entity: Organization: None\|Person: None\|Location: Word1, Word2\|Miscellaneous: Word3" // Text: {text} // Entity: |
| 1-shot | Please identify Organization, Person, Location and Miscellaneous Entity from the given text // Text: AL-AIN , United Arab Emirates 1996-12-06 // Entity: Organization: None\|Person: None\|Location: AL-AIN, United Arab Emirates\|Miscellaneous: None // Text: {text} // Entity:<br><br>Please list all Organization, Person, Location and Miscellaneous Entity in the given text // Text: AL-AIN , United Arab Emirates 1996-12-06 // Entity: Organization: None\|Person: None\|Location: AL-AIN, United Arab Emirates\|Miscellaneous: None // Text: {text} // Entity:<br><br>Extract all Organization, Person, Location and Miscellaneous Entity from the given text // Text: AL-AIN , United Arab Emirates 1996-12-06 // Entity: Organization: None\|Person: None\|Location: AL-AIN, United Arab Emirates\|Miscellaneous: None // Text: {text} // Entity: |

Table 31: 0/1-shot prompts for OntoNotes v5 dataset. The "{text}" should be replaced by sentence and the "{format}" should be replaced by "Entity: Organization: None|Person: Word1|Geo-political Entity: None|Facility: None|Location: Word2|Time: Word3|Cardinal: None|Money: None|Date: None|Percent: None|Language: None|Work of art: None|Nationalities or religious or political groups: Word4, Word5|Quantity: None|Ordinal: None|Product: None|Event: None|Law: None".

| # Shot | Prompts |
| --- | --- |
| zero-shot | Please identify Organization, Person, Geo-political Entity, Facility, Location, Time, Cardinal, Money, Date, Percent, Language, Work of art, Nationalities or religious or political groups, Quantity, Ordinal, Product, Event, Law Entity from the given text, output using the format as '{format}' // Text: {text} // Entity:<br><br>Please list all Organization, Person, Geo-political Entity, Facility, Location, Time, Cardinal, Money, Date, Percent, Language, Work of art, Nationalities or religious or political groups, Quantity, Ordinal, Product, Event, Law Entity in the given text, output using the format as '{format}' // Text: {text} // Entity:<br><br>Extract all Organization, Person, Geo-political Entity, Facility, Location, Time, Cardinal, Money, Date, Percent, Language, Work of art, Nationalities or religious or political groups, Quantity, Ordinal, Product, Event, Law Entity from the given text, output using the format as '{format}' // Text: {text} // Entity: |
| 1-shot | Please identify Organization, Person, Geo-political Entity, Facility, Location, Time, Cardinal, Money, Date, Percent, Language, Work of art, Nationalities or religious or political groups, Quantity, Ordinal, Product, Event, Law Entity from the given text // Text: Graphic by Tsai Chih - pen // Entity: Organization: None|Person: Tsai Chih - pen|Geo-political Entity: None|Facility: None|Location: None|Time: None|Cardinal: None|Money: None|Date: None|Percent: None|Language: None|Work of art: None|Nationalities or religious or political groups: None|Quantity: None|Ordinal: None|Product: None|Event: None|Law: None // Text: {text} // Entity:<br><br>Please list all Organization, Person, Geo-political Entity, Facility, Location, Time, Cardinal, Money, Date, Percent, Language, Work of art, Nationalities or religious or political groups, Quantity, Ordinal, Product, Event, Law Entity in the given text // Text: Graphic by Tsai Chih - pen // Entity: Organization: None|Person: Tsai Chih - pen|Geo-political Entity: None|Facility: None|Location: None|Time: None|Cardinal: None|Money: None|Date: None|Percent: None|Language: None|Work of art: None|Nationalities or religious or political groups: None|Quantity: None|Ordinal: None|Product: None|Event: None|Law: None // Text: {text} // Entity:<br><br>Extract all Organization, Person, Geo-political Entity, Facility, Location, Time, Cardinal, Money, Date, Percent, Language, Work of art, Nationalities or religious or political groups, Quantity, Ordinal, Product, Event, Law Entity from the given text // Text: Graphic by Tsai Chih - pen // Entity: Organization: None|Person: Tsai Chih - pen|Geo-political Entity: None|Facility: None|Location: None|Time: None|Cardinal: None|Money: None|Date: None|Percent: None|Language: None|Work of art: None|Nationalities or religious or political groups: None|Quantity: None|Ordinal: None|Product: None|Event: None|Law: None // Text: {text} // Entity: |

Table 32: 0/1-shot prompts for HONOR dataset. The "{text}" should be replaced by sentence.

| # Shot | Prompts |
|---|---|
| zero-shot | 请识别文本中的所有结束日期、参与人、开始日期、开始时间、开始结束时间、发生地、结束时间、开始结束日期，每个词最多出现在一个类别，使用"结果：结束日期：无，参与人：无，开始日期：词语1，开始时间：词语2，开始结束时间：无，发生地：无，结束时间：词语4，开始结束日期：无"的格式输出// 文本：{text} // 结果：<br><br>请从文本中识别结束日期、参与人、开始日期、开始时间、开始结束时间、发生地、结束时间、开始结束日期并列举出来，每个词最多出现在一个类别，使用"结果：结束日期：无，参与人：无，开始日期：词语1，开始时间：词语2，开始结束时间：无，发生地：无，结束时间：词语4，开始结束日期：无"的格式输出// 文本：{text} // 结果：<br><br>请告诉我给定文本中的结束日期、参与人、开始日期、开始时间、开始结束时间、发生地、结束时间、开始结束日期是什么，每个词最多出现在一个类别，使用"结果：结束日期：无，参与人：无，开始日期：词语1，开始时间：词语2，开始结束时间：无，发生地：无，结束时间：词语4，开始结束日期：无"的格式输出// 文本：{text} // 结果： |
| 1-shot | 请识别文本中的所有结束日期、参与人、开始日期、开始时间、开始结束时间、发生地、结束时间、开始结束日期，每个词最多出现在一个类别// 文本：让他帮我参加一下每天早上的会// 结果：结束日期：无，参与人：无，开始日期：每天，开始时间：早上，开始结束时间：无，发生地：无，结束时间：无，开始结束日期：无// 文本：{text} // 结果：<br><br>请从文本中识别结束日期、参与人、开始日期、开始时间、开始结束时间、发生地、结束时间、开始结束日期并列举出来，每个词最多出现在一个类别// 文本：让他帮我参加一下每天早上的会// 结果：结束日期：无；参与人：无；开始日期：每天；开始时间：早上；开始结束时间：无；发生地：无；结束时间：无；开始结束日期：无// 文本：{text} // 结果：<br><br>请告诉我给定文本中的结束日期、参与人、开始日期、开始时间、开始结束时间、发生地、结束时间、开始结束日期是什么，每个词最多出现在一个类别// 文本：让他帮我参加一下每天早上的会// 结果：结束日期：无；参与人：无；开始日期：每天；开始时间：早上；开始结束时间：无；发生地：无；结束时间：无；开始结束日期：无// 文本：{text} // 结果： |

Table 33: 0/1-shot prompts for MSRANER dataset. The "{text}" should be replaced by sentence.

| # Shot | Prompts |
| --- | --- |
| zero-shot | 请识别文本中的所有人名、地名、组织名，每个词最多出现在一个类别，使用"结果：人名：词语1，词语2;地名：无;组织名：词语3"的格式输出// 文本：{text} // 结果： |
| | 请从给定文本中识别人名、地名、组织名并列举出来，每个词最多出现在一个类别，使用"结果：人名：词语1，词语2;地名：无;组织名：词语3"的格式输出// 文本：{text} // 结果： |
| | 请告诉我给定文本中的人名、地名、组织名是什么，每个词最多出现在一个类别，使用"结果：人名：词语1，词语2;地名：无;组织名：词语3"的格式输出// 文本：{text} // 结果： |
| 1-shot | 请识别文本中的所有人名、地名、组织名，每个词最多出现在一个类别// 文本：中共中央致中国致公党十一大的贺词// 结果：人名：无;地名：无;组织名：中共中央，中国致公党十一大// 文本：{text} // 结果： |
| | 请从给定文本中识别人名、地名、组织名并列举出来，每个词最多出现在一个类别// 文本：中共中央致中国致公党十一大的贺词// 结果：人名：无;地名：无;组织名：中共中央，中国致公党十一大// 文本：{text} // 结果： |
| | 请告诉我给定文本中的人名、地名、组织名是什么，每个词最多出现在一个类别// 文本：中共中央致中国致公党十一大的贺词// 结果：人名：无;地名：无;组织名：中共中央，中国致公党十一大// 文本：{text} // 结果： |

Table 34: 0/1-shot prompts for OntoNote4NER dataset. $ is used as a separator. The "{text}" should be replaced by sentence.

| # Shot | Prompts |
| --- | --- |
| zero-shot | 请识别文本中的所有地缘政治实体、地名、组织机构名、人名，每个词最多出现在一个类别，使用"结果：地缘政治实体$$词语1$词语2$词语3$词语4$$$地名$$无$$$组织名$$无$$$人名$$词语5"的格式输出// 文本：{text} // 结果 |
| | 请从文本中识别地缘政治实体、地名、组织机构名、人名并列举出来，每个词最多出现在一个类别，使用"结果：地缘政治实体$$词语1$词语2$词语3$词语4$$$地名$$无$$$组织名$$无$$$人名$$词语5"的格式输出// 文本：{text} // 结果 |
| | 请告诉我给定文本中的地缘政治实体、地名、组织机构名、人名是什么，每个词最多出现在一个类别，每个词最多出现在一个类别，使用"结果：地缘政治实体$$词语1$词语2$词语3$词语4$$$地名$$无$$$组织名$$无$$$人名$$词语5"的格式输出// 文本：{text} // 结果 |
| 1-shot | 请识别文本中的所有地缘政治实体、地名、组织机构名、人名，每个词最多出现在一个类别// 文本：二次大战日本结束统治后，台湾回归中国，帛琉则成为美国的托管地，并于一九八〇年代开始与我国有了政治上的接触。// 结果：地缘政治实体$$日本$台湾$中国$帛琉$$$地名$$无$$$组织名$$无$$$人名$$无// 文本：{text} // 结果 |
| | 请从文本中识别地缘政治实体、地名、组织机构名、人名并列举出来，每个词最多出现在一个类别// 文本：二次大战日本结束统治后，台湾回归中国，帛琉则成为美国的托管地，并于一九八〇年代开始与我国有了政治上的接触。// 结果：地缘政治实体$$日本$台湾$中国$帛琉$$$地名$$无$$$组织名$$无$$$人名$$无// 文本：{text} // 结果： |
| | 请告诉我给定文本中的地缘政治实体、地名、组织机构名、人名是什么，每个词最多出现在一个类别// 文本：二次大战日本结束统治后，台湾回归中国，帛琉则成为美国的托管地，并于一九八〇年代开始与我国有了政治上的接触。// 结果：地缘政治实体$$日本$台湾$中国$帛琉$$$地名$$无$$$组织名$$无$$$人名$$无// 文本：{text} // 结果： |

Table 35: 0/1-shot prompts for Tacred dataset. The "{subj}" should be replaced by subject word, the "{obj}" should be replaced by object word, and the "{options}" should be replaced by "person and age, no relation, person and title, organization and top members or employees, organization and country of headquarters, person and parents, person and countries of residence, person and children, organization and alternate names, person and charges, person and cities of residence, person and origin, organization and founded by, person and employee of, person and sibling, person and alternate names, organization and website, person and religion, person and state or province of birth, organization and parents, organization and subsidiaries, person and other family, person and state or provinces of residence, organization and members, person and cause of death, organization and member of, organization and number of employees or members, person and country of birth, organization and shareholders, organization and state or province of headquarters, person and city of death, person and date of birth, person and spouse, organization and city of headquarters, person and date of death, person and schools attended, organization and political or religious affiliation, person and country of death, organization and founded, person and state or province of birth, person and city of birth, organization and dissolved".

| # Shot | Prompts |
|---|---|
| zero-shot | '{token}' // In above text, what is the relationship between '{subj}' and '{obj}'? // Options: '{options}' // Answer:<br><br>'{token}' // Determine the relationship between '{subj}' and '{obj}' in above sentence. // Options: '{options}' // Answer:<br><br>'{token}' // Find the relationship between '{subj}' and '{obj}' from above sentence. // Options: '{options}' // Answer: |
| 1-shot | 'Graham , 55 , has maintained his innocence in the killing .' // In above text, what is the relationship between 'Graham' and '55'? // Answer: person and age // '{token}' // In above text, what is the relationship between '{subj}' and '{obj}'? // Options: '{options}' // Answer:<br><br>'Graham , 55 , has maintained his innocence in the killing .' // Determine the relationship between 'Graham' and '55' in above sentence. // Answer: person and age // '{token}' // Determine the relationship between '{subj}' and '{obj}' in above sentence. // Options: '{options}' // Answer:<br><br>'Graham , 55 , has maintained his innocence in the killing .' // Find the relationship between 'Graham' and '55' from above sentence. // Answer: person and age // '{token}' // Find the relationship between '{subj}' and '{obj}' from above sentence. // Options: '{options}' // Answer: |

Table 36: 0/1-shot prompts for MNLI-m dataset. The "{premise}" should be replaced by premise, and the "{hypothesis}" should be replaced by hypothesis.

| # Shot | Prompts |
| --- | --- |
| zero-shot | '{premise}' Based on the previous passage, is it entailment or neutral or contradiction that '{hypothesis}' // Answer: <br><br> Suppose '{premise}' Can we infer that '{hypothesis}'? Please choose one answer: entailment, contradiction, neutral // Answer: <br><br> Given that '{premise}' Therefore, it must be entailment or contradiction or neutral that '{hypothesis}' // Answer: |
| 1-shot | 'He was of two minds, one reveled in the peace of this village.' Based on the previous passage, is it entailment or neutral or contradiction that 'He loved how peaceful the village was.' // Answer: entailment // '{premise}' Based on the previous passage, is it entailment or neutral or contradiction that '{hypothesis}' // Answer: <br><br> Suppose 'He was of two minds, one reveled in the peace of this village.' Can we infer that 'He loved how peaceful the village was.'? Please choose one answer: entailment, contradiction, neutral // Answer: entailment // Suppose '{premise}' Can we infer that '{hypothesis}'? Please choose one answer: entailment, contradiction, neutral // Answer: <br><br> Given that 'He was of two minds, one reveled in the peace of this village.' Therefore, it must be entailment or contradiction or neutral that 'He loved how peaceful the village was.' // Answer: entailment // Given that '{premise}' Therefore, it must be entailment or contradiction or neutral that '{hypothesis}' // Answer: |

Table 37: 0/1-shot prompts for MNLI-mm dataset. The "{premise}" should be replaced by the premise, and the "{hypothesis}" should be replaced by the hypothesis.

| # Shot | Prompts |
| --- | --- |
| zero-shot | '{premise}' Based on the previous passage, is it entailment or neutral or contradiction that '{hypothesis}' // Answer: |
| | Suppose '{premise}' Can we infer that '{hypothesis}'? Please choose one answer: entailment, contradiction, neutral // Answer: |
| | Given that '{premise}' Therefore, it must be entailment or contradiction or neutral that '{hypothesis}' // Answer: |
| 1-shot | 'I'll twist him, sir.' Based on the previous passage, is it entailment or neutral or contradiction that 'I'll make him straight.' // Answer: contradiction // '{premise}' Based on the previous passage, is it entailment or neutral or contradiction that '{hypothesis}' // Answer: |
| | Suppose 'I'll twist him, sir.' Can we infer that 'I'll make him straight.'? Please choose one answer: entailment, contradiction, neutral // Answer: contradiction // Suppose '{premise}' Can we infer that '{hypothesis}'? Please choose one answer: entailment, contradiction, neutral // Answer: |
| | Given that 'I'll twist him, sir.' Therefore, it must be entailment or contradiction or neutral that 'I'll make him straight.' // Answer: contradiction // Given that '{premise}' Therefore, it must be entailment or contradiction or neutral that '{hypothesis}' // Answer:' |

Table 38: 0/1-shot prompts for SNLI dataset. The "{premise}" should be replaced by the premise, and the "{hypothesis}" should be replaced by the hypothesis.

| # Shot | Prompts |
| --- | --- |
| zero-shot | '{premise}' Based on the previous passage, is it entailment or neutral or contradiction that '{hypothesis}' // Answer:<br><br>Suppose '{premise}' Can we infer that '{hypothesis}'? Please choose one answer: entailment, contradiction, neutral // Answer:<br><br>Given that '{premise}' Therefore, it must be entailment or contradiction or neutral that '{hypothesis}' // Answer: |
| 1-shot | Premise:'A person on a horse jumps over a broken down airplane.'. Based on this premise, can we conclude the hypothesis 'A person is training his horse for a competition.' is true? // Options: neutral, contradiction, entailment // Answer: neutral // Premise:'{premise}'. Based on this premise, can we conclude the hypothesis '{hypothesis}' is true? // Options: neutral, contradiction, entailment // Answer:<br><br>Suppose 'A person on a horse jumps over a broken down airplane.' // Can we infer that 'A person is training his horse for a competition.'? // Options: neutral, contradiction, entailment // Answer: neutral // Suppose '{premise}' // Can we infer that '{hypothesis}'? // options:neutral,contradiction,entailment // Answer:<br><br>Given that 'A person on a horse jumps over a broken down airplane.' Therefore, it must be true that 'A person is training his horse for a competition.'? // Options: neutral, contradiction, entailment // Answer: neutral // Given that '{premise}' Therefore, it must be true that '{hypothesis}'? // options:neutral,contradiction,entailment // Answer: |

Table 39: 0/1-shot prompts for WSJ dataset. ￥ is used as a separator. The "{text}" should be replaced by sentence and the "{candidate}" should be replaced by "adjective, plural noun, preposition or conjunction, determiner, singular proper noun, coordinating conjunction, past tense verb, singular or mass noun, wh-determiner, modal, base form verb, wh-adverb, comma, gerund or present partical, to, possessive ending, sentence boundary marker, possessive wh-pronoun, non-3rd person singular present verb, left round bracket, right round bracket, adverb, past participle verb, 3rd person singular present verb, left double quotation mark, right double quotation mark, comparative adverb, monetary values, cardinal number, comparative adjective, particle, personal pronoun, colon character, possessive pronoun, predeterminer, superlative adverb, wh-pronoun, superlative adjective, foreign word, list marker, interjection, existential there, pound symbol, plural proper noun, symbol".

| # Shot | Prompts |
| --- | --- |
| zero-shot | Do part-of-speech task for the given text using the categories in candidate list, output using the format as "Word1￥Categary￥￥Word2￥Categary￥￥Word3￥Category" // Candidate list: {candidate} // Text: {text} // Result:<br><br>Tag the parts of speech in the given text using the categories in candidate list, output using the format as "Word1￥Categary￥￥Word2￥Categary￥￥Word3￥Category" // Candidate list: {candidate} // Text: {text} // Result:<br><br>Label the words in the given text using the categories in candidate list, output using the format as "Word1￥Categary￥￥Word2￥Categary￥￥Word3￥Category" // Candidate list: {candidate} // Text: {text} // Result: |
| 1-shot | Do part-of-speech task for the given text using the categories in candidate list // Candidate list: {candidate} // Text: Few changes were made in the way the markets are regulated . // Result: Few￥adjective￥￥changes￥plural noun￥￥were￥past tense verb￥￥made￥past participle verb￥￥in￥preposition or conjuction￥￥the￥determiner￥￥way￥singular or mass noun￥￥the￥determiner￥￥markets￥plural noun￥￥are￥non-3rd person singular present verb￥￥regulated￥past participle verb￥￥.￥sentence boundary marker // Text: {text} // Result:<br><br>Tag the parts of speech in the given text using the categories in candidate list // Candidate list: {candidate} // Text: Few changes were made in the way the markets are regulated . // Result: Few￥adjective￥￥changes￥plural noun￥￥were￥past tense verb￥￥made￥past participle verb￥￥in￥preposition or conjuction￥￥the￥determiner￥￥way￥singular or mass noun￥￥the￥determiner￥￥markets￥plural noun￥￥are￥non-3rd person singular present verb￥￥regulated￥past participle verb￥￥.￥sentence boundary marker // Text: {text} // Result:<br><br>Label the words in the given text using the categories in candidate list // Candidate list: {candidate} // Text: Few changes were made in the way the markets are regulated . // Result: Few￥adjective￥￥changes￥plural noun￥￥were￥past tense verb￥￥made￥past participle verb￥￥in￥preposition or conjuction￥￥the￥determiner￥￥way￥singular or mass noun￥￥the￥determiner￥￥markets￥plural noun￥￥are￥non-3rd person singular present verb￥￥regulated￥past participle verb￥￥.￥sentence boundary marker // Text: {text} // Result: |

Table 40: 0/1-shot prompts for Daily547 dataset. ¥ is used as a separator. The "{text}" should be replaced by sentence and the "{candidate}" should be replaced by "common noun, pronoun, proper noun, nominal + possessive, proper noun + possessive, verb incl. copula and auxiliaries, adjective, adverb, interjection, determine, pre- or postposition/subordinating conjunction, coordinating conjunction, verb partical, existential there/predeterminers, hashtag, at-mention, discourse marker, URL/email address, emoticon, numeral, punctuation, other, nominal + verbal, proper noun + verbal, X + verbal".

| # Shot | Prompts |
| --- | --- |
| zero-shot | Do part-of-speech task for the given text using the categories in candidate list, output using the format as "Word1¥Categary¥¥Word2¥Categary¥¥Word3¥Category" // Candidate list: {candidate} // Text: {text} // Result:<br><br>Tag the parts of speech in the given text using the categories in candidate list, output using the format as "Word1¥Categary¥¥Word2¥Categary¥¥Word3¥Category" // Candidate list: {candidate} // Text: {text} // Result:<br><br>Label the words in the given text using the categories in candidate list, output using the format as "Word1¥Categary¥¥Word2¥Categary¥¥Word3¥Category" // Candidate list: {candidate} // Text: {text} // Result: |
| 1-shot | Do part-of-speech task for the given text using the categories in candidate list // Candidate list: {candidate} // Text: Bridalplasty ! Love this showww . // Result: Bridalplasty¥proper noun¥¥!¥punctuation¥¥Love¥verb incl. copula and auxiliaries¥¥this¥determine¥¥showww¥common noun¥¥.¥punctuation // Text: {text} // Result:<br><br>Tag the parts of speech in the given text using the categories in candidate list // Candidate list: {candidate} // Text: Bridalplasty ! Love this showww . // Result: Bridalplasty¥proper noun¥¥!¥punctuation¥¥Love¥verb incl. copula and auxiliaries¥¥this¥determine¥¥showww¥common noun¥¥.¥punctuation // Text: {text} // Result:<br><br>Label the words in the given text using the categories in candidate list // Candidate list: {candidate} // Text: Bridalplasty ! Love this showww . // Result: Bridalplasty¥proper noun¥¥!¥punctuation¥¥Love¥verb incl. copula and auxiliaries¥¥this¥determine¥¥showww¥common noun¥¥.¥punctuation // Text: {text} // Result: |

Table 41: 0/1-shot prompts for PKU-SEGPOS dataset. The "{text}" should be replaced by sentence.

| # Shot | Prompts |
| --- | --- |
| zero-shot | 请使用候选集的词性，对给定文本中的每个词语进行标注,使用"词语1_词性/词语2_词性/词语3_词性"的格式输出// 候选集：名词, 时间词, 处所词, 方位词, 数词, 量词, 区别词, 代词, 动词, 形容词, 状态词, 副词, 介词, 连词, 助词, 语气词, 叹词, 拟声词, 成语, 习用语, 简称词, 前接成分, 后接成分, 语素, 非语素字, 标点符号// 文本：{text} // 结果： |
| | 请使用候选集标注给定文本中每个词的词性,使用"词语1_词性/词语2_词性/词语3_词性"的格式输出// 候选集：名词, 时间词, 处所词, 方位词, 数词, 量词, 区别词, 代词, 动词, 形容词, 状态词, 副词, 介词, 连词, 助词, 语气词, 叹词, 拟声词, 成语, 习用语, 简称词, 前接成分, 后接成分, 语素, 非语素字, 标点符号// 文本：{text} // 结果： |
| | 对于给定的文本，请使用候选集标注每个词的词性,使用"词语1_词性/词语2_词性/词语3_词性"的格式输出// 候选集：名词, 时间词, 处所词, 方位词, 数词, 量词, 区别词, 代词, 动词, 形容词, 状态词, 副词, 介词, 连词, 助词, 语气词, 叹词, 拟声词, 成语, 习用语, 简称词, 前接成分, 后接成分, 语素, 非语素字, 标点符号// 文本：{text} // 结果： |
| 1-shot | 请使用候选集的词性，对给定文本中的每个词语进行标注// 候选集：名词, 时间词, 处所词, 方位词, 数词, 量词, 区别词, 代词, 动词, 形容词, 状态词, 副词, 介词, 连词, 助词, 语气词, 叹词, 拟声词, 成语, 习用语, 简称词, 前接成分, 后接成分, 语素, 非语素字, 标点符号// 文本：天津开发区蒸蒸日上。// 结果：天津_名词/开发区_名词/蒸蒸日上_成语/。_标点符号// 文本：{text} // 结果： |
| | 请使用候选集标注给定文本中每个词的词性// 候选集：名词, 时间词, 处所词, 方位词, 数词, 量词, 区别词, 代词, 动词, 形容词, 状态词, 副词, 介词, 连词, 助词, 语气词, 叹词, 拟声词, 成语, 习用语, 简称词, 前接成分, 后接成分, 语素, 非语素字, 标点符号// 文本：天津开发区蒸蒸日上。// 结果：天津_名词/开发区_名词/蒸蒸日上_成语/。_标点符号// 文本：{text} // 结果： |
| | 对于给定的文本，请使用候选集标注每个词的词性// 候选集：名词, 时间词, 处所词, 方位词, 数词, 量词, 区别词, 代词, 动词, 形容词, 状态词, 副词, 介词, 连词, 助词, 语气词, 叹词, 拟声词, 成语, 习用语, 简称词, 前接成分, 后接成分, 语素, 非语素字, 标点符号// 文本：天津开发区蒸蒸日上。// 结果：天津_名词/开发区_名词/蒸蒸日上_成语/。_标点符号// 文本：{text} // 结果： |

Table 42: 0/1-shot prompts for IMDB dataset. The "{sentence}" should be replaced by sentence.

| # Shot | Prompts |
| --- | --- |
| zero-shot | The sentiment expressed for the movie is positive or negative? // '{sentence}' // options:positive,negative // Answer:<br><br>The following movie review expresses what sentiment? positive or negative? // '{sentence}' // options:positive,negative // Answer:<br><br>What is the sentiment expressed by the reviewer for the movie? positive or negative? // '{sentence}' // options:positive,negative // Answer: |
| 1-shot | The sentiment expressed for the movie is positive or negative? // 'The Great Dictator is a beyond-excellent film. Charlie Chaplin succeeds in being both extremely funny and witty and yet at the same time provides a strong statement in his satire against fascism. The anti-Nazi speech by Chaplin at the end, with its values, is one of filmdom's great moments. Throughout this movie, I sensed there was some higher form of intelligence, beyond genuinely intelligent filmmaking, at work.' // options:positive,negative // Answer: positive // '{sentence}' // options:positive,negative // Answer:<br><br>The following movie review expresses what sentiment? positive or negative? // 'The Great Dictator is a beyond-excellent film. Charlie Chaplin succeeds in being both extremely funny and witty and yet at the same time provides a strong statement in his satire against fascism. The anti-Nazi speech by Chaplin at the end, with its values, is one of filmdom's great moments. Throughout this movie, I sensed there was some higher form of intelligence, beyond genuinely intelligent filmmaking, at work.' // Answer: positive // '{sentence}' // options:positive,negative // Answer:<br><br>What is the sentiment expressed by the reviewer for the movie? positive or negative? // 'The Great Dictator is a beyond-excellent film. Charlie Chaplin succeeds in being both extremely funny and witty and yet at the same time provides a strong statement in his satire against fascism. The anti-Nazi speech by Chaplin at the end, with its values, is one of filmdom's great moments. Throughout this movie, I sensed there was some higher form of intelligence, beyond genuinely intelligent filmmaking, at work.' // Answer: positive // '{sentence}' // options:positive,negative // Answer: |

Table 43: 0/1-shot prompts for MRPC dataset. The "{sentence1}" should be replaced by the first sentence, and the "{sentence2}" should be replaced by the second sentence.

| # Shot | Prompts |
| --- | --- |
| zero-shot | Does the sentence '{sentence1}' paraphrase (that is, mean the same thing as) this sentence? '{sentence2}' // Options: yes, no // Answer:<br><br>I want to know whether the following two sentences mean the same thing. '{sentence1}' '{sentence2}' // Options: yes, no // Answer:<br><br>Do the following two sentences mean the same thing? // '{sentence1}' // '{sentence2}' // Options: yes, no // Answer: |
| 1-shot | Does the sentence 'Amrozi accused his brother , whom he called " the witness " , of deliberately distorting his evidence .' paraphrase (that is, mean the same thing as) this sentence? 'Referring to him as only " the witness " , Amrozi accused his brother of deliberately distorting his evidence .' // Options: yes, no // Answer: Yes // Does the sentence '{sentence1}' paraphrase (that is, mean the same thing as) this sentence? '{sentence2}' // Options: yes, no // Answer:<br><br>I want to know whether the following two sentences mean the same thing. 'Amrozi accused his brother , whom he called " the witness " , of deliberately distorting his evidence .' 'Referring to him as only " the witness " , Amrozi accused his brother of deliberately distorting his evidence .' // Options: yes, no // Answer: Yes // I want to know whether the following two sentences mean the same thing. '{sentence1}' '{sentence2}' // Options: yes, no // Answer:<br><br>Do the following two sentences mean the same thing? // 'Amrozi accused his brother , whom he called " the witness " , of deliberately distorting his evidence .' // 'Referring to him as only " the witness " , Amrozi accused his brother of deliberately distorting his evidence .' // Options: yes, no // Answer: Yes // Do the following two sentences mean the same thing? // '{sentence1}' // '{sentence2}' // Options: yes, no // Answer: |

Table 44: 0/1-shot prompts for QQP dataset. The "{question1}" should be replaced by the first question, and the "{question}" should be replaced by the second question.

| # Shot | Prompts |
| --- | --- |
| zero-shot | Can an answer to '{question1}' also be used to answer '{question2}'? // Answer:<br><br>Are the questions '{question1}' and '{question2}' asking the same thing? Yes or no? // Answer:<br><br>I want to know whether the following two questions mean the same thing. '{question1}' '{question2}' Do they? // Options: yes, no // Answer: |
| 1-shot | Can an answer to 'What is the step by step guide to invest in share market in india?' also be used to answer 'What is the step by step guide to invest in share market?'? // Answer: No // Can an answer to '{question1}' also be used to answer '{question2}'? // Answer:<br><br>Are the questions 'What is the step by step guide to invest in share market in india?' and 'What is the step by step guide to invest in share market?' asking the same thing? Yes or no? // Answer: No // Are the questions '{question1}' and '{question2}' asking the same thing? Yes or no? // Answer:<br><br>I want to know whether the following two questions mean the same thing. 'What is the step by step guide to invest in share market in india?' 'What is the step by step guide to invest in share market?' Do they? // Options: yes, no // Answer: No // I want to know whether the following two questions mean the same thing. '{question1}' '{question2}' Do they? // Options: yes, no // Answer: |

Table 45: 0/1-shot prompts for WSC273 dataset. The "{text}" should be replaced by sentence, the "{target 1}" should be replaced by the first target, and the "{target 2}" should be replaced by the second target.

| # Shot | Prompts |
| --- | --- |
| zero-shot | '{text}' // In the previous sentences, does '{target 2}' refer to '{target 1}'? // Options: yes, no // Answer: |
| | '{text}' // Here, does '{target 2}' stand for '{target 1}'? // Options: yes, no // Answer: |
| | '{text}' // In the passage above, can '{target 2}' be replaced by '{target 1}'? // Options: yes, no // Answer: |
| 1-shot | 'the board of aldermen refused the protesters a permit because they feared violence .' // In the previous sentences, does 'they' refer to 'the board of aldermen'? // Answer: yes // '{text}' // In the previous sentences, does '{target 2}' refer to '{target 1}'? // Options: yes, no // Answer: |
| | 'the board of aldermen refused the protesters a permit because they feared violence .' // Here, does 'they' stand for 'the board of aldermen'? // Answer: yes // '{text}' // Here, does '{target 2}' stand for '{target 1}'? // Options: yes, no // Answer: |
| | 'the board of aldermen refused the protesters a permit because they feared violence .' // In the passage above, can 'they' be replaced by 'the board of aldermen'? // Answer: yes // '{text}' // In the passage above, can '{target 2}' be replaced by '{target 1}'? // Options: yes, no // Answer: |